\documentclass[12pt]{article}

\usepackage{newtxtext,newtxmath}

\usepackage{graphicx}

\usepackage[letterpaper,margin=1in]{geometry}

\renewenvironment{abstract}
	{\quotation}
	{\endquotation}

\date{}

\makeatletter
\renewcommand{\fnum@figure}{\textbf{Figure \thefigure}}
\renewcommand{\fnum@table}{\textbf{Table \thetable}}
\makeatother

\usepackage{scicite}

\usepackage{url}

\usepackage{booktabs}%
\usepackage{tabularx}
\usepackage{csquotes}
\usepackage{hyperref}
\usepackage{xspace}
\usepackage{siunitx}
\DeclareSIUnit\bpm{bpm}
\DeclareSIUnit\Unit{Unit}
\newcommand{\autocite}[1]{~\cite{#1}}
\newcommand{\model}[1]{\texttt{#1}}

\newcommand{\varPValuePalmSepsisOxygen}{\num{7.1e-04}\xspace}
\newcommand{\varPValuePalmSepsisPerfusion}{\num{1.1e-01}\xspace}
\newcommand{\varPValuePalmSepsisHaemoglobin}{\num{6.2e-05}\xspace}
\newcommand{\varPValuePalmSepsisWater}{\num{4.5e-10}\xspace}
\newcommand{\varPValuePalmSurvivalOxygen}{\num{6.8e-04}\xspace}
\newcommand{\varPValuePalmSurvivalPerfusion}{\num{2.5e-03}\xspace}
\newcommand{\varPValuePalmSurvivalHaemoglobin}{\num{6.0e-04}\xspace}
\newcommand{\varPValuePalmSurvivalWater}{\num{7.0e-05}\xspace}

\newcommand{\varStatisticsTable}{\begin{tabular}{lllrrrr}
\toprule
site & target & functional parameter & $p$-value & \acs*{dof} & $t$-statistic & \SI{95}{\percent} \acs*{ci} \\
\midrule
palm & sepsis status & oxygen saturation & \num{7.1e-04} & 208 & -3.44 & $[-0.06; -0.02]$ \\
palm & sepsis status & perfusion index & \num{1.1e-01} & 205 & -1.63 & $[-0.03; 0.00]$ \\
palm & sepsis status & haemoglobin index & \num{6.2e-05} & 198 & 4.09 & $[0.03; 0.08]$ \\
palm & sepsis status & water index & \num{4.5e-10} & 222 & 6.53 & $[0.03; 0.06]$ \\
palm & survival status & oxygen saturation & \num{6.8e-04} & 79 & -3.54 & $[-0.09; -0.02]$ \\
palm & survival status & perfusion index & \num{2.5e-03} & 82 & -3.12 & $[-0.06; -0.01]$ \\
palm & survival status & haemoglobin index & \num{6.0e-04} & 81 & 3.57 & $[0.03; 0.09]$ \\
palm & survival status & water index & \num{7.0e-05} & 93 & 4.16 & $[0.02; 0.05]$ \\
finger & sepsis status & oxygen saturation & \num{1.4e-04} & 176 & -3.89 & $[-0.07; -0.02]$ \\
finger & sepsis status & perfusion index & \num{1.5e-03} & 196 & -3.22 & $[-0.06; -0.02]$ \\
finger & sepsis status & haemoglobin index & \num{4.4e-07} & 205 & 5.22 & $[0.05; 0.10]$ \\
finger & sepsis status & water index & \num{1.2e-01} & 194 & 1.56 & $[-0.00; 0.02]$ \\
finger & survival status & oxygen saturation & \num{3.7e-04} & 75 & -3.73 & $[-0.10; -0.03]$ \\
finger & survival status & perfusion index & \num{5.4e-04} & 79 & -3.61 & $[-0.09; -0.03]$ \\
finger & survival status & haemoglobin index & \num{4.6e-04} & 81 & 3.65 & $[0.03; 0.11]$ \\
finger & survival status & water index & \num{5.6e-01} & 84 & -0.59 & $[-0.02; 0.01]$ \\
\bottomrule
\end{tabular}
\xspace}
\newcommand{\varTotalSubjects}{508\xspace}

\newcommand{\varFocusRelativeAbdominalFocus}{\SI{53}{\percent}\xspace}
\newcommand{\varFocusRelativeRespiratoryFocus}{\SI{17}{\percent}\xspace}
\newcommand{\varFocusRelativeMutipleFocus}{\SI{8}{\percent}\xspace}
\newcommand{\varFocusRelativeUnkownFocus}{\SI{14}{\percent}\xspace}
\newcommand{\varFocusRelativeSkinSoftTissueFocus}{\SI{5}{\percent}\xspace}
\newcommand{\varFocusRelativeGenitourinaryFocus}{\SI{3}{\percent}\xspace}

\newcommand{\varMortalityRateSepsisSepticShock}{\SI{49}{\percent} ($24/49$)\xspace}
\newcommand{\varTotalSepsisTask}{437\xspace}
\newcommand{\varTotalExclusionSepsisTask}{71\xspace}

\newcommand{\varTotalSepsis}{129\xspace}
\newcommand{\varRelativeSepsis}{\SI{30}{\percent}\xspace}
\newcommand{\varMortalityRateSepsis}{\SI{27}{\percent} ($35/129$)\xspace}
\newcommand{\varTotalNosepsis}{308\xspace}
\newcommand{\varRelativeNosepsis}{\SI{70}{\percent}\xspace}
\newcommand{\varMortalityRateNosepsis}{\SI{6}{\percent} ($18/308$)\xspace}
\newcommand{\varTotalSurvivalTask}{483\xspace}

\newcommand{\varTotalNonsurvivor}{68\xspace}
\newcommand{\varRelativeNonsurvivor}{\SI{14}{\percent}\xspace}
\newcommand{\varDescriptiveTableOneHrs}{\begin{tabularx}{\textwidth}{>{\raggedright}p{2.5cm}XXXX}
\toprule
attribute & no sepsis & sepsis & non survivor & survivor \\
\midrule
number of subjects & 308 & 129 & 68 & 415 \\
\multicolumn{5}{c}{\textbf{demographics}} \\
age & \num{6.2e+01} (\num{1.5e+01}) & \num{6.6e+01} (\num{1.4e+01}) & \num{6.9e+01} (\num{1.5e+01}) & \num{6.3e+01} (\num{1.4e+01}) \\
sex & 220 male\newline88 female & 90 male\newline39 female & 41 male\newline27 female & 299 male\newline116 female \\
weight [\si{\kg}] & \num{8.2e+01} (\num{2.0e+01}) & \num{8.2e+01} (\num{2.6e+01}) & \num{7.5e+01} (\num{2.3e+01}) & \num{8.2e+01} (\num{2.1e+01}) \\
type of weight measurement & 245 estimated\newline53 measured & 100 estimated\newline16 measured & 52 estimated\newline8 measured & 331 estimated\newline68 measured \\
\multicolumn{5}{c}{\textbf{vital signs}} \\
heart frequency [\si{\bpm}] & \num{8.2e+01} (\num{1.7e+01}) & \num{9.9e+01} (\num{2.1e+01}) & \num{9.8e+01} (\num{2.4e+01}) & \num{8.6e+01} (\num{1.9e+01}) \\
sinusrhythm [\si{\percent}] & \num{79} & \num{74} & \num{60} & \num{78} \\
MAP [\si{\mmHg}] & \num{8.1e+01} (\num{1.4e+01}) & \num{7.6e+01} (\num{1.3e+01}) & \num{7.7e+01} (\num{1.3e+01}) & \num{8.0e+01} (\num{1.4e+01}) \\
systolic blood pressure & \num{1.2e+02} (\num{2.3e+01}) & \num{1.2e+02} (\num{1.9e+01}) & \num{1.2e+02} (\num{2.3e+01}) & \num{1.2e+02} (\num{2.3e+01}) \\
temperature [\si{\degreeCelsius}] & \num{3.7e+01} (\num{6.7e-01}) & \num{3.7e+01} (\num{1.1e+00}) & \num{3.7e+01} (\num{1.1e+00}) & \num{3.7e+01} (\num{7.5e-01}) \\
SpO2 [\si{\percent}] & \num{9.7e+01} (\num{2.2e+00}) & \num{9.7e+01} (\num{4.0e+00}) & \num{9.6e+01} (\num{5.1e+00}) & \num{9.7e+01} (\num{2.3e+00}) \\
\multicolumn{5}{c}{\textbf{BGA measurements}} \\
pCO2 [\si{\mmHg}] & \num{3.9e+01} (\num{5.8e+00}) & \num{4.4e+01} (\num{9.8e+00}) & \num{4.3e+01} (\num{9.8e+00}) & \num{4.0e+01} (\num{7.0e+00}) \\
pO2 [\si{\mmHg}] & \num{9.8e+01} (\num{3.4e+01}) & \num{1.0e+02} (\num{2.5e+01}) & \num{1.0e+02} (\num{2.5e+01}) & \num{9.9e+01} (\num{3.2e+01}) \\
sO2 [\si{\percent}] & \num{9.7e+01} (\num{1.6e+00}) & \num{9.6e+01} (\num{2.8e+00}) & \num{9.6e+01} (\num{3.5e+00}) & \num{9.7e+01} (\num{1.6e+00}) \\
Hb (BGA) [\si{\g\per\deci\litre}] & \num{9.7e+00} (\num{1.7e+00}) & \num{9.4e+00} (\num{1.7e+00}) & \num{9.5e+00} (\num{1.6e+00}) & \num{9.5e+00} (\num{1.7e+00}) \\
lactate [\si{\mg\per\deci\litre}] & \num{1.6e+01} (\num{1.4e+01}) & \num{2.7e+01} (\num{3.4e+01}) & \num{4.6e+01} (\num{5.3e+01}) & \num{1.5e+01} (\num{1.1e+01}) \\
pH & \num{7.4e+00} (\num{5.8e-02}) & \num{7.4e+00} (\num{8.8e-02}) & \num{7.4e+00} (\num{1.0e-01}) & \num{7.4e+00} (\num{6.5e-02}) \\
type BGA & 274 arterial\newline7 venous & 104 arterial\newline1 venous & 56 arterial & 358 arterial\newline10 venous \\
\multicolumn{5}{c}{\textbf{organ replacement therapies}} \\
renal replacement therapy [\si{\percent}] & \num{4} & \num{20} & \num{28} & \num{7} \\
ECMO [\si{\percent}] & \num{1} & \num{2} & \num{3} & \num{1} \\
impella [\si{\percent}] & \num{0} & \num{1} & \num{4} & \num{0} \\
liver replacement therapy [\si{\percent}] & \num{1} & \num{2} & \num{4} & \num{0} \\
\multicolumn{5}{c}{\textbf{ventilation parameters}} \\
invasive ventilation [\si{\percent}] & \num{48} & \num{95} & \num{93} & \num{59} \\
ventilation [\si{\percent}] & \num{23} & \num{80} & \num{78} & \num{34} \\
APRV [\si{\percent}] & \num{0} & \num{2} & \num{3} & \num{0} \\
FiO2 [\si{\percent}] & \num{3.2e+01} (\num{1.0e+01}) & \num{4.4e+01} (\num{1.8e+01}) & \num{4.4e+01} (\num{1.7e+01}) & \num{3.4e+01} (\num{1.3e+01}) \\
PEEP [\si{\milli\bar}] & \num{7.0e+00} (\num{2.3e+00}) & \num{8.9e+00} (\num{3.2e+00}) & \num{8.3e+00} (\num{3.3e+00}) & \num{8.1e+00} (\num{2.9e+00}) \\
P-peak [\si{\milli\bar}] & \num{2.0e+01} (\num{5.5e+00}) & \num{2.1e+01} (\num{6.1e+00}) & \num{2.2e+01} (\num{5.7e+00}) & \num{2.0e+01} (\num{5.8e+00}) \\
respiratory frequency [\si{\minute\tothe{-1}}] & \num{1.7e+01} (\num{4.4e+00}) & \num{1.8e+01} (\num{5.3e+00}) & \num{1.7e+01} (\num{5.6e+00}) & \num{1.7e+01} (\num{5.0e+00}) \\
\multicolumn{5}{c}{\textbf{dose of administered vasopressors and inotropes}} \\
noradrenaline dose [\si{\ug\per\kg\per\minute}] & \num{4.4e-02} (\num{9.4e-02}) & \num{2.6e-01} (\num{2.6e-01}) & \num{2.7e-01} (\num{3.0e-01}) & \num{7.7e-02} (\num{1.4e-01}) \\
adrenaline dose [\si{\ug\per\kg\per\minute}] & \num{9.2e-04} (\num{1.1e-02}) & \num{3.7e-03} (\num{2.3e-02}) & \num{8.6e-03} (\num{3.2e-02}) & \num{7.0e-04} (\num{1.0e-02}) \\
vasopressin dose [\si{\Unit\per\kg\per\minute}] & \num{3.8e-06} (\num{3.2e-05}) & \num{5.4e-05} (\num{1.4e-04}) & \num{5.2e-05} (\num{1.1e-04}) & \num{1.2e-05} (\num{7.3e-05}) \\
dobutamine dose [\si{\ug\per\kg\per\minute}] & \num{2.0e-01} (\num{9.3e-01}) & \num{6.1e-01} (\num{1.9e+00}) & \num{1.1e+00} (\num{2.4e+00}) & \num{2.7e-01} (\num{1.2e+00}) \\
\bottomrule
\end{tabularx}\xspace}
\newcommand{\varDescriptiveTableTenHrs}{\begin{tabularx}{\textwidth}{>{\raggedright}p{2.5cm}XXXX}
\toprule
attribute & no sepsis & sepsis & non survivor & survivor \\
\midrule
creatinine [\si{\mg\per\deci\litre}] & \num{1.3e+00} (\num{1.1e+00}) & \num{1.9e+00} (\num{1.5e+00}) & \num{1.7e+00} (\num{9.7e-01}) & \num{1.5e+00} (\num{1.3e+00}) \\
GFR [\si{\ml\per\minute}] & \num{7.3e+01} (\num{3.6e+01}) & \num{4.9e+01} (\num{3.4e+01}) & \num{4.6e+01} (\num{2.9e+01}) & \num{6.7e+01} (\num{3.7e+01}) \\
LDH [\si{\Unit\per\litre}] & \num{5.4e+02} (\num{7.8e+02}) & \num{6.8e+02} (\num{1.6e+03}) & \num{1.3e+03} (\num{2.3e+03}) & \num{4.8e+02} (\num{6.7e+02}) \\
bilirubin [\si{\mg\per\deci\litre}] & \num{1.9e+00} (\num{2.4e+00}) & \num{2.4e+00} (\num{3.5e+00}) & \num{2.7e+00} (\num{3.7e+00}) & \num{1.9e+00} (\num{2.4e+00}) \\
CRP [\si{\mg\per\litre}] & \num{6.6e+01} (\num{7.4e+01}) & \num{2.0e+02} (\num{1.1e+02}) & \num{1.2e+02} (\num{9.5e+01}) & \num{1.1e+02} (\num{1.1e+02}) \\
leukocytes [\si{\nano\litre\tothe{-1}}] & \num{1.1e+01} (\num{5.0e+00}) & \num{1.6e+01} (\num{1.1e+01}) & \num{1.5e+01} (\num{9.7e+00}) & \num{1.3e+01} (\num{7.3e+00}) \\
Hb (lab) [\si{\g\per\deci\litre}] & \num{9.9e+00} (\num{1.9e+00}) & \num{9.8e+00} (\num{1.8e+00}) & \num{9.6e+00} (\num{1.6e+00}) & \num{9.8e+00} (\num{1.9e+00}) \\
platelets [\si{\nano\litre\tothe{-1}}] & \num{1.6e+02} (\num{8.3e+01}) & \num{2.1e+02} (\num{1.4e+02}) & \num{1.8e+02} (\num{1.2e+02}) & \num{1.8e+02} (\num{1.1e+02}) \\
hematocrit [\si{\percent}] & \num{2.9e-01} (\num{5.3e-02}) & \num{3.0e-01} (\num{5.5e-02}) & \num{2.9e-01} (\num{5.1e-02}) & \num{2.9e-01} (\num{5.4e-02}) \\
sodium [\si{\milli\mole\per\liter}] & \num{1.4e+02} (\num{4.3e+00}) & \num{1.4e+02} (\num{6.0e+00}) & \num{1.4e+02} (\num{5.9e+00}) & \num{1.4e+02} (\num{4.9e+00}) \\
potassium [\si{\milli\mole\per\liter}] & \num{4.5e+00} (\num{5.3e-01}) & \num{4.7e+00} (\num{6.2e-01}) & \num{4.7e+00} (\num{6.8e-01}) & \num{4.5e+00} (\num{5.4e-01}) \\
PCT [\si{\nano\g\per\ml}] & \num{1.9e+00} (\num{7.7e+00}) & \num{5.2e+01} (\num{1.6e+02}) & \num{2.3e+01} (\num{6.3e+01}) & \num{1.6e+01} (\num{9.6e+01}) \\
\bottomrule
\end{tabularx}
\xspace}
\newcommand{\varSepsisPerformancePalmHSI}[1][\ac]{0.80 (\SI{95}{\percent} #1{ci} [0.76; 0.84])\xspace}

\newcommand{\varSepsisPerformanceFingerHSI}[1][\ac]{0.72 (\SI{95}{\percent} #1{ci} [0.67; 0.78])\xspace}

\newcommand{\varSurvivalPerformancePalmHSI}[1][\ac]{0.72 (\SI{95}{\percent} #1{ci} [0.65; 0.79])\xspace}

\newcommand{\varSurvivalPerformanceFingerHSI}[1][\ac]{0.66 (\SI{95}{\percent} #1{ci} [0.59; 0.73])\xspace}

\newcommand{\varMaxRGBHSIImprovement}{\SI{23}{\percent}\xspace}
\newcommand{\varTotalFeaturesOneHrs}{33\xspace}
\newcommand{\varSepsisHSIPlusClinicalDataPerformanceOneHrsPlusOne}[1][\ac]{0.87 (\SI{95}{\percent} #1{ci} [0.83; 0.90])\xspace}

\newcommand{\varSepsisHSIPlusClinicalDataPerformanceOneHrsPlusAll}[1][\ac]{0.90 (\SI{95}{\percent} #1{ci} [0.87; 0.92])\xspace}

\newcommand{\varTotalFeaturesTenHrs}{45\xspace}

\newcommand{\varSepsisHSIPlusClinicalDataPerformanceTenHrsPlusAll}[1][\ac]{0.94 (\SI{95}{\percent} #1{ci} [0.92; 0.96])\xspace}

\newcommand{\varSurvivalHSIPlusClinicalDataPerformanceOneHrsPlusThree}[1][\ac]{0.80 (\SI{95}{\percent} #1{ci} [0.74; 0.85])\xspace}

\newcommand{\varSurvivalHSIPlusClinicalDataPerformanceOneHrsPlusAll}[1][\ac]{0.82 (\SI{95}{\percent} #1{ci} [0.76; 0.88])\xspace}

\newcommand{\varSurvivalHSIPlusClinicalDataPerformanceTenHrsPlusAll}[1][\ac]{0.83 (\SI{95}{\percent} #1{ci} [0.78; 0.88])\xspace}

\newcommand{\varTotalFeaturesLab}{12\xspace}
\newcommand{\varSepticShockPerformancePalmHSI}[1][\ac]{0.66 (\SI{95}{\percent} #1{ci} [0.57; 0.75])\xspace}
\newcommand{\varSepticShockPerformanceFingerHSI}[1][\ac]{0.62 (\SI{95}{\percent} #1{ci} [0.51; 0.71])\xspace}
\newcommand{\varSepsisPerformancePalmFirstpatchHSISepsisBias}[1][\ac]{0.91 (\SI{95}{\percent} #1{ci} [0.85; 0.96])\xspace}
\newcommand{\varSepsisPerformancePalmOodpatchHSISepsisBias}[1][\ac]{0.73 (\SI{95}{\percent} #1{ci} [0.69; 0.78])\xspace}

\newif\ifhighlight
\highlightfalse

\usepackage[dvipsnames]{xcolor}
\ifhighlight \newcommand{\highlight}[1]{{\color{Orchid} #1}} \else \newcommand{\highlight}[1]{#1} \fi
\ifhighlight \newcommand{\strike}[1]{\st{#1}} \else \newcommand{\strike}[1]{} \fi

\usepackage{acro}
\newcommand{\acrocolor}[1]{\textcolor{black}{#1}}
\acsetup{make-links=true,format=\acrocolor}

\DeclareAcronym{hsi}{short=HSI,long=hyperspectral imaging}
\DeclareAcronym{sd}{short=SD,long=standard deviation}
\DeclareAcronym{ci}{short=CI,long=confidence interval}
\DeclareAcronym{icu}{short=ICU,short-plural=\acrocolor{s},long=intensive care unit}
\DeclareAcronym{vis}{short=VIS,long=vasoactive inotropic score}
\DeclareAcronym{news}{short=NEWS,long=National Early Warning Score}
\DeclareAcronym{ehr}{short=EHR,short-plural=\acrocolor{s},long=electronic health record}
\DeclareAcronym{lmics}{short=LMICs,long=low- and middle-income countries}
\DeclareAcronym{crt}{short=CRT,long=capillary refill time}
\DeclareAcronym{sms}{short=SMS,long=skin mottling score}
\DeclareAcronym{iqr}{short=IQR,long=interquartile range}
\DeclareAcronym{roc}{short=ROC,long=receiver operating characteristic}
\DeclareAcronym{auroc}{short=AUROC,long=area under the receiver operating characteristic curve}
\DeclareAcronym{crp}{short=CRP,long=C-reactive protein}
\DeclareAcronym{pct}{short=PCT,long=procalcitonin}
\DeclareAcronym{sofa}{short=SOFA,long=Sequential Organ Failure Assessment}
\DeclareAcronym{map}{short=MAP,long=mean arterial pressure}
\DeclareAcronym{bga}{short=BGA,long=blood gas analysis}
\DeclareAcronym{pco2}{short=pCO2,long=carbon dioxide partial pressure}
\DeclareAcronym{po2}{short=pO2,long=oxygen partial pressure}
\DeclareAcronym{so2}{short=sO2,long=oxygen saturation}
\DeclareAcronym{hb}{short=Hb,long=haemoglobin concentration}
\DeclareAcronym{aprv}{short=APRV,long=airway pressure release ventilation}
\DeclareAcronym{pcv}{short=PCV,long=pressure-controlled ventilation}
\DeclareAcronym{peep}{short=PEEP,long=positive endexpiratory pressure}
\DeclareAcronym{ppeak}{short=P-peak,long=peak inspiratory pressure}
\DeclareAcronym{fio2}{short=FiO2,long=fraction of inspired oxygen}
\DeclareAcronym{phigh}{short=P-high,long=high airway pressure}
\DeclareAcronym{thigh}{short=T-high,long=duration of high airway pressure}
\DeclareAcronym{plow}{short=P-low,long=low airway pressure}
\DeclareAcronym{ecmo}{short=ECMO,long=extracorporeal membrane oxygenation}
\DeclareAcronym{mars}{short=MARS,long=Molecular Adsorbents Recirculating System}
\DeclareAcronym{qsofa}{short=qSOFA,long=quick Sequential Organ Failure Assessment}
\DeclareAcronym{sirs}{short=SIRS,long=Systemic Inflammatory Response Syndrome}
\DeclareAcronym{apache}{short=APACHE,long=Acute Physiology and Chronic Health Evaluation}
\DeclareAcronym{dof}{short=DOF,long=degrees of freedom}
\DeclareAcronym{rgb}{short=RGB,long=red-green-blue}
\DeclareAcronym{led}{short=LED,long=light-emitting diode}
\DeclareAcronym{rfe}{short=RFE,long=recursive feature elimination}
\DeclareAcronym{spo2}{short=SpO2,long=pulse oxymetrical oxygen saturation}
\DeclareAcronym{gfr}{short=\acrocolor{GFR},long=glomerular filtration rate}
\DeclareAcronym{ldh}{short=\acrocolor{LDH},long=lactate dehydrogenase}
\DeclareAcronym{cnn}{short=\acrocolor{CNN},long=convolutional neural network}

\def\scititle{
	AI-powered skin spectral imaging enables instant sepsis diagnosis and outcome prediction in critically ill patients
}
\title{\bfseries \boldmath \scititle}

\author{
	Silvia~Seidlitz$^{1,2,3,4\ast}$,
	Katharina~Hölzl$^{1,5}$,
    Ayca~von~Garrel$^{1,5}$,
    Jan~Sellner$^{1,2,3,4}$,\and
    Stephan~Katzenschlager$^{5}$,
    Tobias~Hölle$^{5}$,
    Dania~Fischer$^{5}$,
    Maik~von~der~Forst$^{5}$,\and
    Felix~C.~F.~Schmitt$^{5}$,
    Alexander~Studier-Fischer$^{6,7,8,9}$,
    Markus~A.~Weigand$^{5\dagger}$,\and
    Lena~Maier-Hein$^{1,2,3,4,10\ast\dagger}$,
    Maximilian~Dietrich$^{5\ast\dagger}$
    \\
	\small$^{1}$Division of Intelligent Medical Systems (IMSY), German Cancer Research Center (DKFZ), Heidelberg, Germany.\and
	\small$^{2}$Helmholtz Information and Data Science School for Health, Heidelberg/Karlsruhe, Germany.\and
    \small$^{3}$Faculty of Mathematics and Computer Science, Heidelberg University, Heidelberg, Germany.\and
    \small$^{4}$National Center for Tumor Diseases (NCT), NCT Heidelberg, a partnership between DKFZ and university\and \small medical center Heidelberg.\and
    \small$^{5}$Heidelberg University, Medical Faculty, Department of Anesthesiology, Heidelberg University Hospital,\and \small Heidelberg, Germany.\and
    \small$^{6}$Department of General, Visceral, and Transplantation Surgery, Heidelberg University Hospital, Heidelberg, Germany.\and
    \small$^{7}$Department of Urology and Urosurgery, Medical Faculty of the University of Heidelberg, \and
    \small University Medical Center Mannheim, Mannheim, Germany.\and
    \small$^{8}$Division of Intelligent Systems and Robotics in Urology (ISRU), German Cancer Research Center (DKFZ), \and \small Heidelberg, Germany.\and
    \small$^{9}$DKFZ Hector Cancer Institute at the University Medical Center Mannheim, Mannheim, Germany.\and
    \small$^{10}$Medical Faculty, Heidelberg University, Heidelberg, Germany.\and
	\small$^\ast$Corresponding author. Email: s.seidlitz@dkfz-heidelberg.de (S.S.); l.maier-hein@dkfz-heidelberg.de (L.M.-H.); \and \small maximilian.dietrich@med.uni-heidelberg.de (M.D.)\and
	\small$^\dagger$These authors contributed equally to this work.
}

\begin{document} 

\maketitle

\begin{abstract} \bfseries \boldmath
With sepsis remaining a leading cause of mortality, early identification of septic patients and those at high risk of death is a challenge of high socioeconomic importance. Given the potential of  \ac{hsi} to monitor microcirculatory alterations, we propose a deep learning approach to automated sepsis diagnosis and mortality prediction using a single  \ac{hsi} cube acquired within seconds. In a prospective observational study, we collected \ac{hsi} data from the palms and fingers of over 480 \ac{icu} patients. Neural networks applied to \ac{hsi} measurements predicted sepsis and mortality with an \ac{auroc} of 0.80 and 0.72, respectively. Performance improved substantially with additional clinical data, reaching \acp{auroc} of 0.94 for sepsis and 0.83 for mortality. We conclude that deep learning-based \ac{hsi} analysis enables rapid and non-invasive prediction of sepsis and mortality, with potential clinical value for enhancing diagnosis and treatment.
\end{abstract}


\paragraph*{Teaser} AI-driven hyperspectral imaging can rapidly and non-invasively diagnose sepsis and predict mortality in critically ill patients.

\section*{Introduction}

Sepsis is defined as a life-threatening organ dysfunction resulting from a dysregulated host response to infection \autocite{singer_sepsis3_2016}. It represents a leading cause of mortality and critical illness worldwide, accounting for \SI{19.7}{\percent} of global deaths in 2017 \autocite{rudd_burden_sepsis_2020}. As the clinical diagnosis of sepsis relies on the presence of organ dysfunction, only patients in advanced stages of the sepsis syndrome are typically identified \autocite{singer_sepsis3_2016}. The resulting delay in sepsis diagnosis is critical as the risk of mortality escalates with each hour of treatment delay due to irreversible organ damage \autocite{ferrer_empiric_2014}. Conversely, patients incorrectly presumed to have sepsis are often treated unnecessarily with antibiotics, which carry risks ranging from mild side effects to severe complications, while simultaneously contributing to the development of multidrug-resistant organisms \autocite{cunha_antibiotic_2001,vincent_annual_update_2023}. A critical aspect of sepsis management is the early and accurate diagnosis prior to the onset of persistent organ dysfunction. This task is complicated by the nonspecific signs and symptoms of the sepsis syndrome, along with the complex, heterogeneous, and not yet fully understood sepsis pathophysiology \autocite{henning_absence_2017}. A particular challenge lies in distinguishing between septic and non-septic critically ill patients in the \ac{icu} due to the higher baseline illness severity and frequent organ failure from both septic and non-septic inflammation \autocite{choi2022mortality}.

Beyond the early identification of septic patients, the early and accurate identification of \ac{icu} patients at high risk of death is crucial. This is because it can substantially improve individual patient outcomes by enabling the timely implementation of appropriate interventions, thereby enhancing patient care \autocite{kumar2022mortality}. Moreover, it has the potential to improve the overall efficiency and effectiveness of critical care delivery. This could be achieved through an optimised allocation of limited resources, informed decisions regarding palliative care, and a deeper understanding of the factors that influence patient outcomes \autocite{iwase2022prediction, kumar2022mortality}.

Over the past decades, considerable research efforts have focused on identifying biomarkers for sepsis diagnosis and mortality prediction, with over 250 molecules proposed as potential diagnostic or prognostic markers. However, to date, no single biomarker has demonstrated outstanding sensitivity and specificity for detecting sepsis and predicting mortality \autocite{pierrakos2020biomarkers}.

More recently, studies have investigated the use of machine learning to predict sepsis and mortality based on high-dimensional clinical data extracted from \acp{ehr} \autocite{komorowski_sepsis_2022}. Despite promising performance metrics reported in research studies \autocite{fleuren2020machine, islam2023machine}, the clinical translation of \ac{ehr}-based sepsis and mortality prediction models faces substantial challenges. \ac{ehr} data, which are primarily collected for the purpose of clinical documentation and billing, lack standardization and contain inaccuracies and biases \autocite{sauer2022leveraging}. These factors can lead to limited generalizability on external datasets, as demonstrated for several \ac{ehr}-based sepsis prediction models \autocite{wong2021external, moor2023predicting}. Furthermore, while \ac{ehr} adoption is widespread in high-income countries, it lags in \ac{lmics}, where \SI{85}{\percent} of sepsis cases occur \autocite{rudd_burden_sepsis_2020,woldemariam2023adoption}.

In recent years, imaging methods such as sublingual microscopy, laser Doppler flowmetry, laser speckle contrast imaging, near-infrared spectroscopy, and \ac{hsi} have revealed that microcirculatory dysfunction, characterised by local zones of hypoxia \autocite{walley1996heterogeneity, ostergaard2015microcirculatory}, develops early during sepsis \autocite{raia2022endothelial} and is a key driver of organ failure and poor outcomes \autocite{de2013microcirculatory, trzeciak2007early}. We therefore hypothesise that \ac{hsi} could enable automated sepsis diagnosis and mortality prediction in \ac{icu} patients by monitoring microcirculatory dysfunction and edema formation. The key strengths of \ac{hsi} include its mobile, non-invasive, rapid, objective, cost-effective, and standardised assessments. Unlike other imaging modalities capable of monitoring microcirculation, medical device-graded \ac{hsi} systems have begun to emerge, paving the way for \ac{hsi} to become a routine clinical tool \autocite{dietrich_bedside_2021, wang2024research}. 
While recent initial work on \ac{hsi}-based sepsis diagnosis performed by ourselves \autocite{dietrich2021machine} and others in parallel \autocite{kohnke2024proof} showed promising performance, all studies conducted so far come with the major limitation that sepsis patients were compared to healthy volunteers or selectively chosen cohorts, such as patients undergoing pancreatic surgeries \autocite{lacis2019hybrid, kazune2019relationship, dietrich_bedside_2021}. Hence, the proposed algorithms are at high risk of shortcut learning due to confounders such as substantial age gaps and differences in comorbidities and therapies between septic patients and non-septic controls \autocite{dietrich2021machine}. Consequently, they are unlikely to generalise well to realistic clinical applications, such as automated sepsis diagnosis in critically ill \ac{icu} patients.

\begin{figure*}[h]
    \centering
    \includegraphics[width=\linewidth]{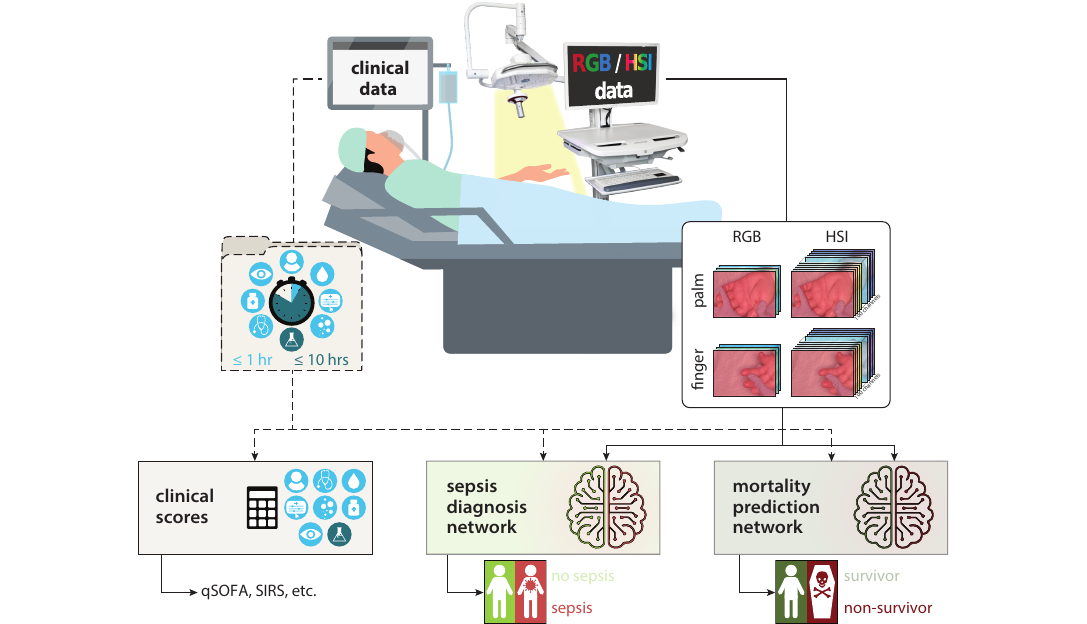}
    \caption{\textbf{We explore \acf*{hsi} for automated, non-invasive and rapid sepsis diagnosis and mortality prediction.} In a prospective study of over 480 \acf*{icu} patients, we collected \acs*{hsi} and \acf*{rgb} images of the palm and annular finger, and clinical data. Deep learning accurately predicts sepsis and mortality from \acs*{hsi} data, with improved performance when combined with clinical data. Our method outperforms widely used clinical biomarkers and scores such as the \acf*{qsofa} score and the \acf*{sirs} criteria.}
    \label{fig:overview}
\end{figure*}

In summary, despite extensive research efforts, robust biomarkers for early sepsis diagnosis and mortality prediction are still lacking. In this article, we close this important gap by presenting the first analysis of deep learning-based \ac{hsi} analysis for automated, non-invasive, and rapid diagnosis of sepsis and prediction of mortality among \ac{icu} patients. Based on a prospective study involving over 480 patients, representing, to the best of our knowledge, the largest medical \ac{hsi} dataset to date, we investigate the following research questions (cf. Figure~\ref{fig:overview}): 

\begin{enumerate}
    \item Is an automated, non-invasive and rapid diagnosis of sepsis and prediction of mortality among \ac{icu} patients feasible with deep learning-based \ac{hsi} analysis? What is the optimal measurement site? Does \ac{hsi} data provide advantages over conventional \ac{rgb} imaging and tissue parameter images derived from \ac{hsi} in terms of classification performance?
    \item Can we further boost the diagnostic and predictive performance by adding structured clinical data?
    \item Does our method outperform widely available clinical scores and biomarkers?
\end{enumerate}

\section*{Results}

Using the medical device-graded \ac{hsi} system Tivita\textsuperscript{\textregistered} 2.0 Surgery Edition (Diaspective Vision, Am Salzhaff, Germany), we collected spectral imaging data from the skin of all patients admitted to the interdisciplinary surgical \ac{icu} at the University Hospital Heidelberg. All adult patients admitted between October 24, 2022, and December 15, 2023, were included, resulting in data from \varTotalSubjects patients.

Of these \varTotalSubjects patients, the sepsis status could not be determined for \varTotalExclusionSepsisTask patients. The sepsis diagnosis cohort is thus composed of the remaining \varTotalSepsisTask patients, of which \varTotalSepsis (\varRelativeSepsis) were diagnosed with sepsis, while \varTotalNosepsis (\varRelativeNosepsis) were not. Among septic patients, the majority (\varFocusRelativeAbdominalFocus) had an abdominal focus, followed by \varFocusRelativeRespiratoryFocus with a respiratory focus, \varFocusRelativeSkinSoftTissueFocus with a skin or soft tissue focus, and \varFocusRelativeGenitourinaryFocus with a genitourinary focus. Additionally, \varFocusRelativeMutipleFocus of the septic patients had multiple foci, while in \varFocusRelativeUnkownFocus the focus of infection remained unknown.

Successful follow-up on 30-day mortality after \ac{icu} admission was achieved for \varTotalSurvivalTask out of the initial \varTotalSubjects patients. These patients constitute the mortality prediction cohort, of which \varTotalNonsurvivor (\varRelativeNonsurvivor) died within 30 days of admission. The mortality rate was higher among patients with sepsis and septic shock at the time of admission, at \varMortalityRateSepsis and \varMortalityRateSepsisSepticShock, respectively, compared to \varMortalityRateNosepsis for those without sepsis at admission.

The palm and annular finger were chosen as measurement sites due to their easy accessibility and low melanin content. Characteristic tissue spectra for septic vs. non-septic patients and survivors vs. non-survivors are available in figure~\ref{fig:spectra}.

\ac{hsi} captures tissue reflectance spectra, which are influenced by chromophores like haemoglobin and water within the tissue. Consequently, functional tissue parameter indices can be approximated from \ac{hsi} data according to the formulas presented in \autocite{holmer_hyperspectral_2018}, including oxygen saturation (the fraction of oxygen-saturated haemoglobin relative to total haemoglobin), perfusion index (a composite measure of perfusion targeting deeper tissue layers), haemoglobin index (indicative of the amount of haemoglobin in the tissue microcirculation), and water index (reflecting the water content in tissue) \autocite{holmer_hyperspectral_2018, kulcke_compact_2018}. Furthermore, an integrated \ac{rgb} sensor captured \ac{rgb} images alongside the \ac{hsi} data, allowing for a direct comparison between \ac{hsi} - a novel imaging modality with enhanced spectral information - and the more widely used, cost-effective, and faster \ac{rgb} imaging.

\subsection*{\ac{hsi} can rapidly and non-invasively diagnose sepsis and predict mortality}

\begin{figure*}[h]
    \centering
    \includegraphics[width=\linewidth]{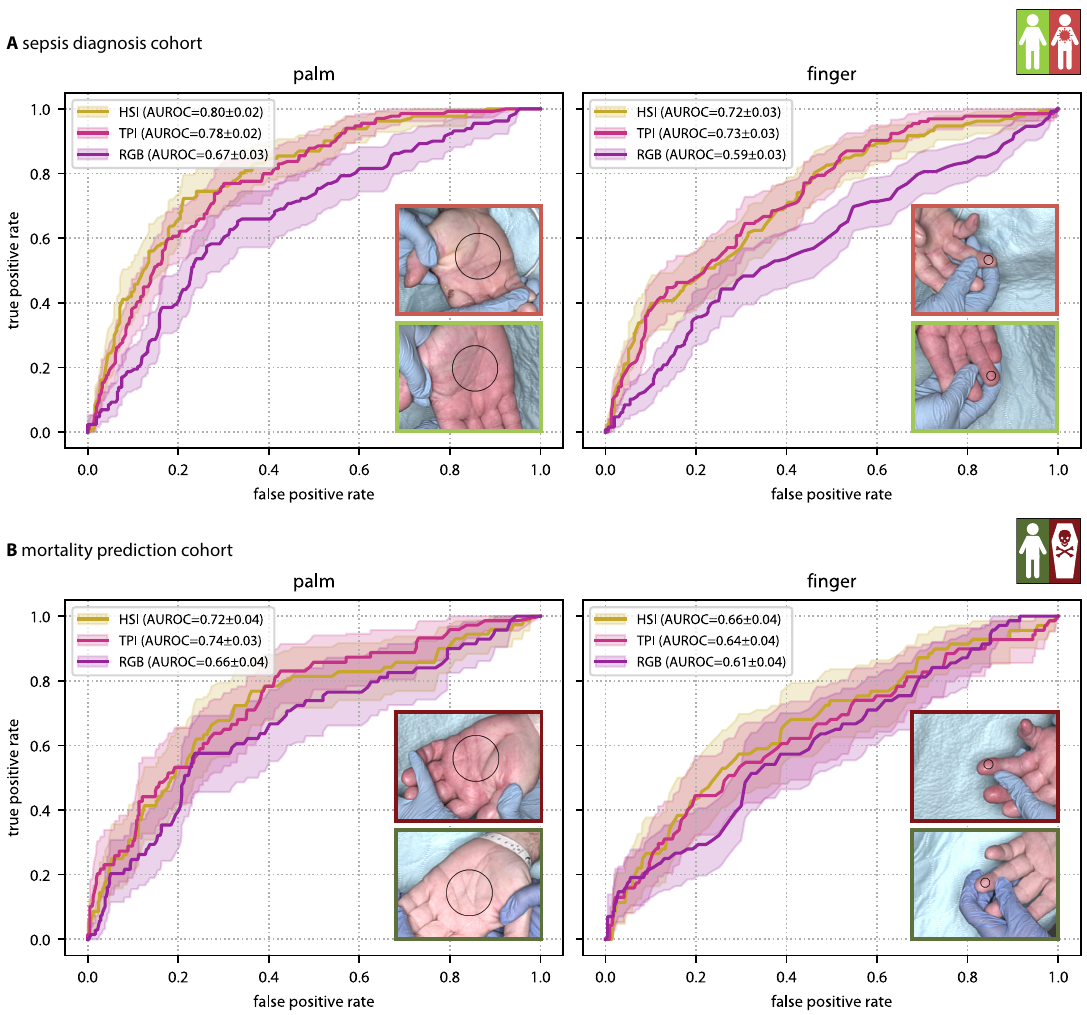}
    \caption{\textbf{\Acf*{hsi} can rapidly and non-invasively diagnose sepsis and predict mortality.} \Acfp*{roc} are shown for sepsis diagnosis \textbf{(A)} and mortality prediction \textbf{(B)} models based on \acs*{hsi} data (gold), stacked tissue parameter images (TPI, pink) and \acf*{rgb} (violet) data of the palm (left) and annular finger (right).The shaded areas denote the \SI{95}{\percent} confidence interval across 1000 bootstrap samples, and mean and standard deviation of the \acf*{auroc} are reported in the legend. Sample images of a septic (light red box) and non-septic (light green box) patient, as well as a survivor (dark green) and non-survivor (dark red) are included on the bottom right, with the circle denoting the annotated skin region.}
    \label{fig:HSIvsRGB}
\end{figure*}

As shown in Figure~\ref{fig:HSIvsRGB}, deep learning-based diagnosis of sepsis from the palm was achievable with an \ac{auroc} of \varSepsisPerformancePalmHSI, while the finger measurements yielded an \ac{auroc} of \varSepsisPerformanceFingerHSI. Also for mortality prediction, the palm measurement site yielded better classification performance (\ac{auroc} \varSurvivalPerformancePalmHSI), compared to the finger measurements (\ac{auroc} \varSurvivalPerformanceFingerHSI). Combining both measurement sites did not yield substantial performance improvements that would justify the added complexity and effort of acquiring two \ac{hsi} measurements instead of one (cf. figure~\ref{fig:performance_combined}).

\ac{hsi} demonstrated superior diagnostic performance compared to conventional \ac{rgb} imaging, with up to a \varMaxRGBHSIImprovement improvement. The performance of models based on \ac{hsi} data and those using tissue parameter images derived from \ac{hsi} data was similar, suggesting that tissue parameter images effectively capture information relevant to sepsis diagnosis and mortality prediction.

\subsection*{Septic patients and non-survivors have decreased palm tissue oxygen saturation at increased tissue water and haemoglobin content}

\begin{figure*}[h]
    \centering
    \includegraphics[width=\linewidth]{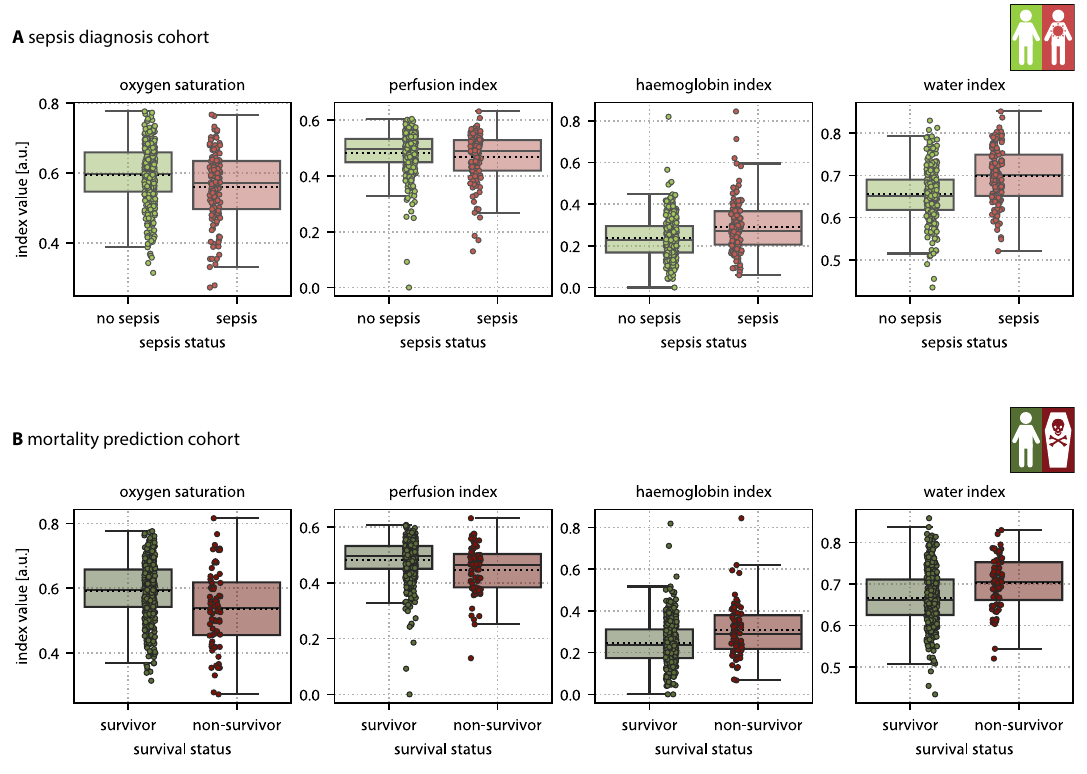}
    \caption{\textbf{Septic patients and non-survivors possess significantly lower palm tissue oxygen saturation, and higher tissue haemoglobin and water index.} The subfigures show the distribution of the functional parameters oxygen saturation, perfusion index, haemoglobin index and water index, derived from \acl*{hsi} palm measurements, for septic and non-septic patients (\textbf{A}), and survivors and non-survivors (\textbf{B}). The boxes denote the quartiles of the distribution with the whiskers extending up to 1.5 times the interquartile range, and the median and mean drawn as solid and dashed lines, respectively. Each dot represents one patient. Tissue parameter index distributions for the measurement site finger are available in figure~\ref{fig:tpi_finger}.}
    \label{fig:parameter_distributions}
\end{figure*}

Distributions of the functional tissue parameter indices oxygen saturation, perfusion index, haemoglobin index and water index for septic and non-septic patients, as well as survivors and non-survivors are shown in Figure~\ref{fig:parameter_distributions} for the measurement site palm. In septic patients, oxygen saturation was significantly lower compared to non-septic patients ($p$ = \varPValuePalmSepsisOxygen), while haemoglobin and water index were significantly higher ($p$ = \varPValuePalmSepsisHaemoglobin and $p$ = \varPValuePalmSepsisWater, respectively). The perfusion index did not show a significant difference ($p$ = \varPValuePalmSepsisPerfusion).  In non-survivors, perfusion index and oxygen saturation were significantly lower ($p$ = \varPValuePalmSurvivalPerfusion and $p$ = \varPValuePalmSurvivalOxygen, respectively) compared to survivors, and haemoglobin and water index were significantly higher ($p$ = \varPValuePalmSurvivalHaemoglobin and $p$ = \varPValuePalmSurvivalWater, respectively). More details on the statistical tests are available in table~\ref{tab:statistical_testing}, and the tissue parameter index distributions for the measurement site finger are illustrated in figure~\ref{fig:tpi_finger}.

\subsection*{Structured clinical data boosts the classification performance}

Structured clinical data were collected alongside the \ac{hsi} data, including demographics, vital signs, blood gas analysis measurements, therapy details (usage of organ replacement therapies, ventilation parameters, and dose of administered vasopressors and inotropes), and laboratory results. A total of \varTotalFeaturesTenHrs clinical parameters were recorded, with \varTotalFeaturesOneHrs usually available within one hour of admission and the \varTotalFeaturesLab laboratory parameters usually available within ten hours of admission. Descriptive statistics of the clinical parameters are available in table~\ref{tab:descriptive_statistics_1hour} and table~\ref{tab:descriptive_statistics_10hours}.

\begin{figure*}
    \centering
    \includegraphics[width=0.85\linewidth]{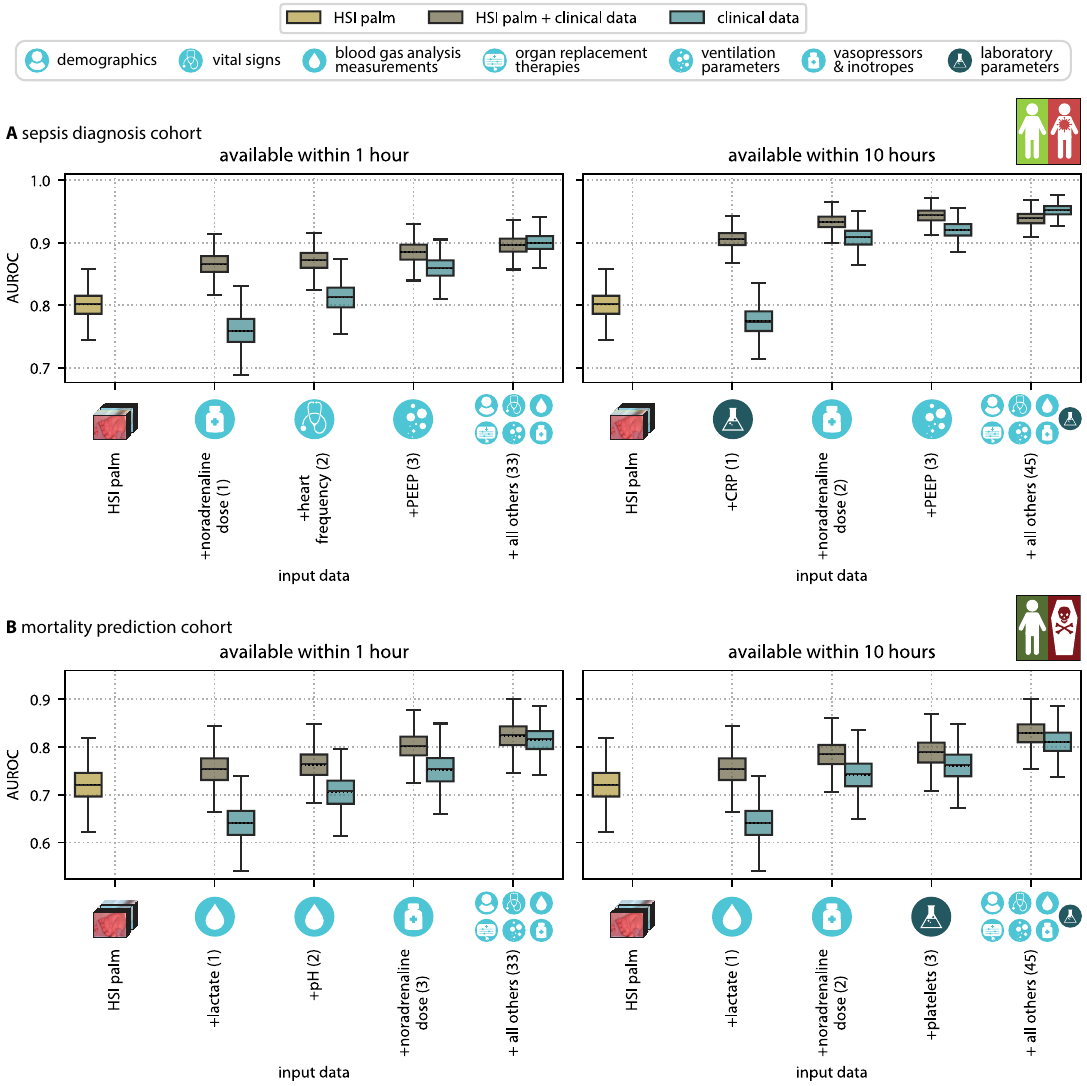}
    \caption{\textbf{Adding clinical data boosts the sepsis diagnosis and mortality prediction performance.} The performance of sepsis diagnosis (\textbf{A}) and mortality prediction (\textbf{B}) using \acf*{hsi} data of the palm (\model{HSI palm} model, gold), a combination of \acs*{hsi} and clinical data (\model{HSI palm + clinical data} model, bronze), and clinical data alone (\model{clinical data} model, blue) is shown, categorised by data availability within one hour (left) and ten hours (right) from admission to the intensive care unit. Within the subplots, the performance of the \model{HSI palm} model is compared to \model{HSI palm + clinical data} and \model{clinical data} models that incorporate - from left to right - the most important, two most important, three most important or all clinical data features available within the specified timeframe of one hour or ten hours after \acl*{icu} admission. The number of clinical data features used in the model is indicated in brackets. The ranking of the clinical data features according to feature importance was derived from the \model{clinical data} model through \acl*{rfe} \autocite{guyon2002gene} starting from the complete set of available clinical data at the given time point. Each box plot represents the quartiles of the \acf*{auroc} distribution across 1000 bootstrap samples, with whiskers extending up to 1.5 times the interquartile range. The median and mean are drawn as solid and dashed lines, respectively.}
    \label{fig:adding_clinical_data}
\end{figure*}

Incorporating all clinical data available within the first hour of \ac{icu} admission alongside \ac{hsi} data of the palm in the \model{HSI palm + clinical data} model improved sepsis diagnosis performance from an \ac{auroc} of \varSepsisPerformancePalmHSI to \varSepsisHSIPlusClinicalDataPerformanceOneHrsPlusAll (cf. Figure~\ref{fig:adding_clinical_data}). The performance further increased to \varSepsisHSIPlusClinicalDataPerformanceTenHrsPlusAll when laboratory values, available within ten hours post-admission, were also included. Although a random forest model using comprehensive clinical data alone, referred to as \model{clinical data} model, performed slightly better on the full set of clinical data, combining \ac{hsi} with clinical data performed substantially better when only limited clinical data were available, a common scenario in emergency settings, outpatient health care and \ac{lmics}. We ranked the importance of clinical data features using \ac{rfe} \autocite{guyon2002gene} with the \model{clinical data} model, beginning with the full set of clinical data available within the specified timeframe of either one hour or ten hours after \ac{icu} admission. An overview of feature importance is presented in figure~\ref{fig:feature_importance_1hour} for clinical data available within one hour and in figure~\ref{fig:feature_importance_10hours} for clinical data available within ten hours. As depicted in Figure~\ref{fig:adding_clinical_data}, sequentially adding clinical data features in the order of their importance revealed that the \model{HSI palm + clinical data} model already achieved an \ac{auroc} of \varSepsisHSIPlusClinicalDataPerformanceOneHrsPlusOne by combining a single clinical parameter immediately available at bedside, namely the administered noradrenaline dose, with \ac{hsi} data.

Combining \ac{hsi} with clinical data also boosted the mortality prediction performance. The \ac{auroc} improved from \varSurvivalPerformancePalmHSI to \varSurvivalHSIPlusClinicalDataPerformanceOneHrsPlusAll when including all clinical data available within the first hour of admission, and further to \varSurvivalHSIPlusClinicalDataPerformanceTenHrsPlusAll when incorporating all clinical data from the first ten hours of admission. When clinical data features were sequentially added in order of their importance, the \model{HSI palm + clinical data} model consistently outperformed the \model{clinical data} model, with the performance advantage being most pronounced when using a smaller number of clinical data features. The three most important clinical data available within one hour from admission were lactate, pH and noradrenaline dose, which achieved an \ac{auroc} of \varSurvivalHSIPlusClinicalDataPerformanceOneHrsPlusThree in combination with \ac{hsi} data of the palm.

\subsection*{Our \ac{hsi}-based classification models surpass clinical biomarkers and scores for sepsis diagnosis and mortality prediction}

We compared our \model{HSI palm + clinical data} models using the complete set of clinical data available within one hour or ten hours after \ac{icu} admission, as well as our \model{HSI palm} models, with commonly used clinical biomarkers and scores for sepsis diagnosis and mortality prediction. 

Rapidly available bedside scores for sepsis diagnosis include the \ac{sms} \autocite{ait2011mottling} and \ac{crt} \autocite{pan2020microcirculation}, both of which depend on visual skin assessment. Additionally, we compared our sepsis diagnosis models to the \ac{qsofa} score \autocite{singer_sepsis3_2016} and the \ac{news} \autocite{rcop2012national}, which are based on vital signs and cognitive function. Among the biomarkers and scores available within ten hours from admission, we compared our models against the inflammatory biomarkers \ac{pct} and \ac{crp}. Additionally, we evaluated them against the \ac{sirs} criteria, previously used for sepsis diagnosis \autocite{bone1992definitions}, and the \ac{sofa} score \autocite{singer_sepsis3_2016}, which is a key component of the current Sepsis-3 definition.

Commonly used clinical biomarkers and scores for assessing disease severity and risk of mortality include the \ac{vis} \autocite{gaies2010vasoactive}, the \ac{sofa} score \autocite{singer_sepsis3_2016}, and the \ac{apache} II score \autocite{knaus1985apache}. The \ac{vis}, which quantifies haemodynamic support based on vasopressor and inotrope doses, is available within one hour of admission. Within ten hours, the \ac{sofa} score, assessing organ dysfunction, and the \ac{apache} II score, measuring disease severity, are available. These scores are derived from various vital signs, laboratory parameters, and patient and therapy characteristics, typically using the most abnormal readings within the last 24 hours. To compare our \model{HSI + clinical data} models with clinical scores, we employed modified versions of the \ac{sofa} and \ac{apache} II scores, which were based on the most current data rather than the worst values over 24 hours, ensuring that values were available on the day of admission.

\begin{figure*}
    \centering
    \includegraphics[width=0.85\linewidth]{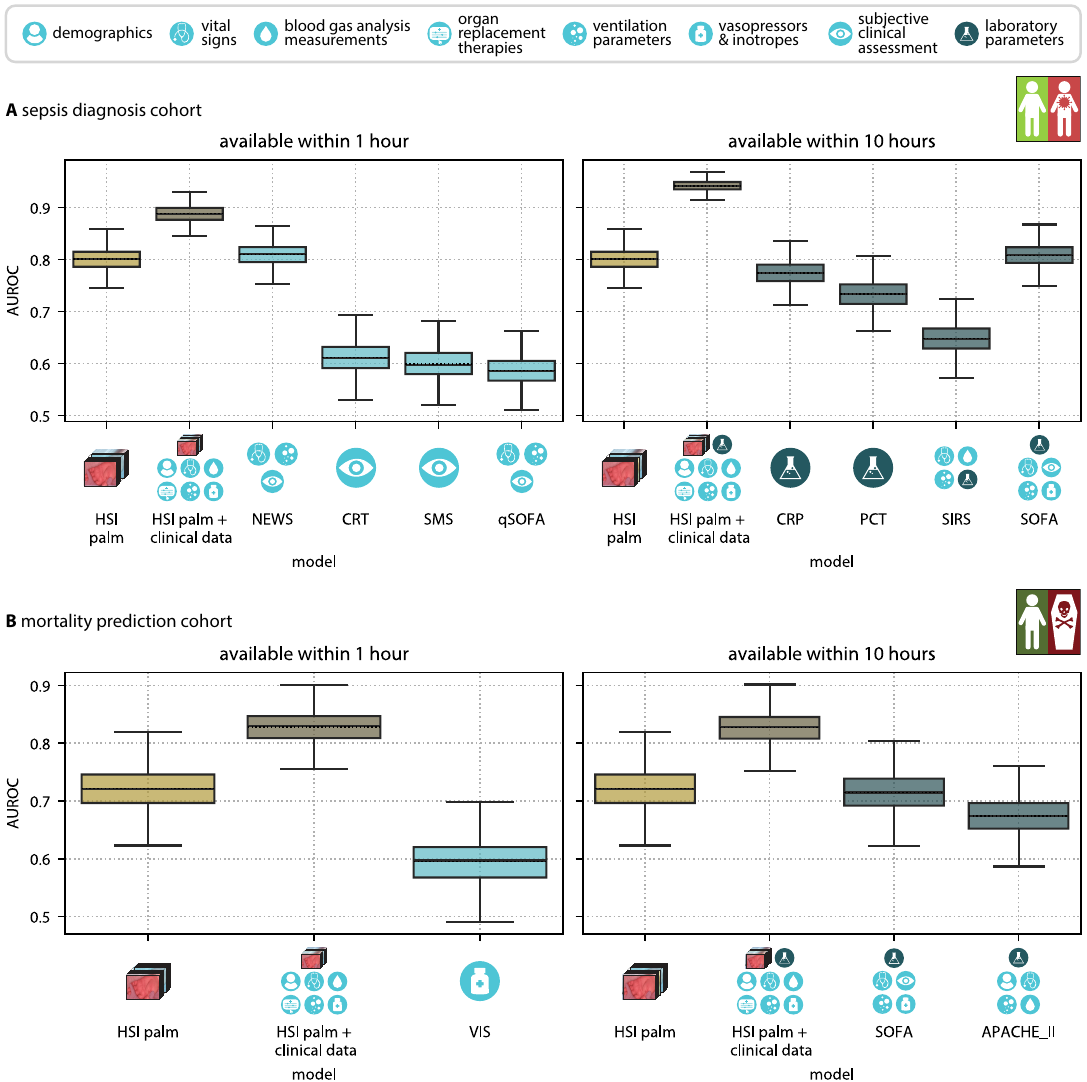}
    \caption{\textbf{Our \model{HSI + clinical data} models outperform widely used clinical biomarkers and scores for sepsis diagnosis and mortality prediction.} Comparison of the \acf*{auroc} for deep learning-based sepsis diagnosis (\textbf{A}) and mortality prediction (\textbf{B}) using \acf*{hsi} data of the palm (\model{HSI palm} model, gold) and a combination of \acs*{hsi} data and the entire set of clinical data available within one hour (left) and ten hours (right) from admission to the intensive care unit (\model{HSI palm + clinical data} model, bronze) against clinical biomarkers and scores (blue). For data available within one hour of \acl*{icu} admission, the comparison includes \acf*{news}, \acf*{crt}, \acf*{sms}, \acf*{qsofa} score, and \acf*{vis}. For data available within ten hours of admission, the comparison includes \acf*{crp}, \acf*{pct}, \acf*{sirs} criteria, \acf*{sofa} score, and \acf*{apache} II score. Each box plot displays the quartiles of the \acs*{auroc} distribution across 1000 bootstrap samples, with whiskers extending up to 1.5 times the interquartile range. The median and mean are represented by solid and dashed lines, respectively.}
    \label{fig:HSIvsScores}
\end{figure*}

As shown in Figure~\ref{fig:HSIvsScores}, our \model{HSI + clinical data} models outperformed all clinical biomarkers and scores for both sepsis diagnosis and mortality prediction.

\section*{Discussion}

In this study, we addressed the critical need for reliable biomarkers to identify septic patients and those at high risk of mortality. We are the first to demonstrate that automated, non-invasive, and rapid diagnosis of sepsis and prediction of mortality among \ac{icu} patients is feasible using deep learning-based \ac{hsi} analysis. Based on the --- to the best of our knowledge --- largest \ac{hsi} patient cohort to date, we derived the following key findings:

\begin{enumerate}
    \item \ac{hsi}-based prediction: Both sepsis and mortality can be predicted from \ac{hsi} data with high accuracy using deep learning. Septic patients and non-survivors have significantly lower tissue oxygen saturation, and higher tissue haemoglobin and water content than non-septic patients and non-survivors. Predictions from \ac{hsi} measurements of the palm are superior to those from the annular finger.
    \item Combination with structured clinical data: Incorporating structured clinical data enhances classification performance, yielding an \ac{auroc} of up to 0.94 and 0.83 for sepsis diagnosis and mortality prediction, respectively.
    \item Comparison to clinical biomarkers and scores: Our \model{HSI + clinical data} models outperform widely available clinical biomarkers and scores that were suggested for sepsis diagnosis and mortality prediction.
\end{enumerate}

\subsection*{Strengths and limitations of our \ac{hsi}-based prediction}

We believe the primary strengths of our \ac{hsi}-based sepsis diagnosis and mortality prediction are its objectivity, non-invasiveness, cost-effectiveness and speed, as predictions can be obtained from a single \ac{hsi} cube acquired at the bedside within seconds. 
\highlight{Given these advantages, our method could be applied as a screening tool across the entire cohort of critically ill \ac{icu} patients, enabling the objective and timely identification of those at high risk for sepsis and mortality. This, in turn, could support the rapid initiation of further diagnostic evaluations and therapeutic interventions.}
Additionally, \ac{hsi} systems enable mobile measurements, and could thus be performed in various hospital wards, such as the emergency department, or even in ambulances. Although \ac{hsi} systems are not yet widely used clinically, they have evolved rapidly over the past two decades from custom research prototypes to medically certified systems, like the one used in this study \autocite{hsi_applications_yoon2022}. Manufacturers such as imec (Leuven, Belgium) and HAIC (Hanover, Germany) are currently focusing on developing more compact, real-time \ac{hsi} devices and scaling production to achieve high-volume, low-cost availability.

We acknowledge that our classification models based on \ac{hsi} data alone may not be sufficiently accurate as a standalone diagnostic and prognostic tool. However, we believe \ac{hsi} has high potential as a pre-screening tool to identify patients for which time-consuming and costly tests (e.g. laboratory measurements) and extensive monitoring should be performed. This is particularly advantageous in resource-limited settings, such as \ac{lmics}, where approximately half of critical care interventions are delivered outside of the \ac{icu} \autocite{bartlett2023critical}, and in situations requiring immediate decisions, such as emergency treatment. 

We showed that substantial performance improvements are possible by integrating a few clinical parameters available at the bedside. For instance, including the administered noradrenaline dose as an additional input improved the \ac{auroc} for sepsis prediction from \varSepsisPerformancePalmHSI to \varSepsisHSIPlusClinicalDataPerformanceOneHrsPlusOne. We want to emphasise, however, that incorporating clinical data may introduce biases and limit generalisability. For example, treatment choices such as the administered noradrenaline dose depend on the implementation of clinical guidelines, which are subject to variation over time and across health systems.

While we showed that the \model{\ac{hsi} + clinical data} models largely outperformed widely used clinical biomarkers and scores and achieved excellent sepsis diagnosis and mortality prediction performance, another limitation of models that require clinical data is that prospectively collecting clinical data requires substantial labour. We decided against the less labour-intensive export of \ac{ehr} data, as several clinical parameters are not reliably recorded in the \ac{ehr}, which would lead to inaccuracies. Additionally, many clinical parameters, such as vital signs and ventilation parameters, were stored in the \ac{ehr} at a poor temporal resolution, failing to accurately reflect the patient's status at the time of the \ac{hsi} measurement. Given the labour-intensive collection of clinical data, minimising the number of clinical parameters required for prediction is advantageous. Our results demonstrate that our \model{HSI + clinical data} models outperform \model{clinical data} models for both sepsis diagnosis and mortality prediction when only few clinical parameters are available.

\subsection*{Comparison to the state of the art}

In 2021, we pioneered machine learning-based sepsis diagnosis using \ac{hsi}. While our framework was able to differentiate sepsis patients from a control group comprising healthy volunteers and patients undergoing pancreatic surgery with high performance (\ac{auroc} of \varSepsisPerformancePalmFirstpatchHSISepsisBias), we identified several potential sources of bias, including differences in age, comorbidities, and therapies between septic and non-septic patients, as well as imaging-related factors such as variations in hardware and acquisition protocols. We concluded that these biases may have inflated the algorithm's performance and could limit its generalisability to real-world clinical settings, such as automated sepsis diagnosis in critically ill \ac{icu} patients \autocite{dietrich2021machine}. In the spirit of good scientific practise, we therefore never submitted our work to peer review but chose to design this new prospective study for automated sepsis diagnosis and mortality prediction specifically in \ac{icu} patients. With our new carefully designed dataset, we achieved lower performance compared to our original work, despite using the same machine learning framework. Moreover, the performance of our model trained on the potentially biased dataset dropped drastically to an \ac{auroc} of \varSepsisPerformancePalmOodpatchHSISepsisBias when applied to the new data. These two facts taken together underscore our hypothesis that prior \ac{hsi} studies \autocite{lacis2019hybrid, kazune2019relationship, dietrich_bedside_2021,dietrich2021machine,kohnke2024proof}, which compared sepsis patients with healthy volunteers or selectively chosen cohorts, have limited relevance for accurately assessing the practical feasibility of automated sepsis diagnosis in real-world clinical settings.

\subsection*{Future work}

A key limitation of our study is that all data were collected from a single surgical \ac{icu} in Germany. As expected for this setting, the majority of septic patients had an abdominal focus, while other infection sites, such as respiratory (\SI{17}{\percent}) and genitourinary (\SI{3}{\percent}), were less common. Furthermore, the management of critically ill patients varies across clinical sites. While some sites manage critically ill patients in emergency wards before transferring them to the \ac{icu}, at our site (Heidelberg University Hospital, Germany), critically ill patients --- whether newly admitted from the emergency ward or those with postoperative complications from the general ward --- are immediately transferred to the \ac{icu}. As a result, septic patients in our \ac{icu} cohort may be in the earlier stages of the disease compared to those in \ac{icu} cohorts at other clinical sites. Given these variations in \ac{icu} populations, external validation is necessary to assess the generalisability of our models across diverse \ac{icu} settings and clinical sites with different patient populations.

Given the key strengths of our \ac{hsi}-based classification models, which enable a rapid, non-invasive, cost-effective, and mobile assessment of sepsis diagnosis and mortality, investigating their performance in resource-constrained and time-critical settings, such as ambulances, emergency wards, and \ac{lmics}, is a promising future direction. Furthermore, it would be valuable to explore whether \ac{hsi}, in addition to detecting septic patients, has the potential to identify individuals earlier in the disease progression, hours or even days before the onset of organ dysfunction. Additionally, since an estimated \SI{40}{\percent} of sepsis cases in 2017 occurred in children under 5 years old \autocite{rudd_burden_sepsis_2020}, expanding the cohort to include infants is of interest.

Furthermore, while our observational study identified potential use cases and demonstrated high accuracy in automated \ac{hsi}-based sepsis diagnosis and mortality prediction, future interventional studies are needed to assess the clinical effectiveness of implementing an automated sepsis and mortality alert system based on our algorithms in the management of critically ill patients. Such studies should compare the system to the standard of care, evaluating its impact on key clinical outcomes, including reductions in mortality, morbidity, and length of hospital stay. To date, only few studies have systematically investigated the clinical effectiveness of automated sepsis and mortality alert systems \autocite{Zhang2022}.

While we consider our single time-point measurements advantageous for enabling immediate diagnosis and low resource requirements, future studies collecting longitudinal \ac{hsi} data could expand the potential of \ac{hsi}. Longitudinal data could improve the understanding of disease progression by identifying features associated with clinical improvement or deterioration.

Besides using \ac{hsi} for disease diagnosis and prognosis, \ac{hsi} holds the potential to support novel therapeutic strategies by continuously assessing tissue microcirculation, evaluating treatment effects, and guiding interventions. In the example of shock therapy, the current therapeutic target is macrohaemodynamic stabilisation (e.g., maintaining normative arterial blood pressure). However, critically ill patients, particularly those with sepsis, often experience a loss of haemodynamic coherence, resulting in dissociation between macro- and microcirculation \autocite{ince2015hemodynamic}. While no other widely available clinical data can capture the spatial distribution of tissue microcirculation, \ac{hsi} offers a unique opportunity for real-time monitoring. While our primary objective was to identify septic patients and those at risk of mortality, we performed an initial experiment investigating whether differentiation between sepsis and septic shock is feasible with \ac{hsi}. As demonstrated in figure~\ref{fig:shock_performance}, a shock classification model achieved an \ac{auroc} of \varSepticShockPerformancePalmHSI at the palm and \varSepticShockPerformanceFingerHSI at the fingers. Future research should further investigate the role of \ac{hsi} in guiding therapy, not only in shock but also in septic patients and those at high risk of mortality, to assess its potential for improving patient outcomes.

\subsection*{Conclusion}

In this study, we addressed the critical need for reliable biomarkers to identify septic patients and those at high risk of mortality. We are the first to investigate the potential of \ac{hsi} for sepsis diagnosis and mortality prediction in \ac{icu} patients, based on a prospective study with the largest \ac{hsi} patient cohort to date, involving over 480 patients. Our proposed \model{HSI} models demonstrated high predictive performance, which improved further when combined with minimal clinical data. They outperformed widely used clinical biomarkers and scores. Key strengths of \ac{hsi}-based predictions include their rapid, non-invasive, cost-effective, and mobile measurements, making them promising candidates for various clinical settings, including resource-limited scenarios (e.g., \ac{lmics}) and time-critical situations (e.g., ambulances, emergency wards). Beyond their proven benefit in sepsis diagnosis and mortality prediction, \ac{hsi}-based microcirculatory monitoring could also offer novel therapeutic strategies and enhance understanding of disease progression. Our code and pretrained models will be made publicly available in our GitHub repository \href{https://github.com/IMSY-DKFZ/htc}{https://github.com/IMSY-DKFZ/htc} and Zenodo \href{https://doi.org/10.5281/zenodo.6577614}{https://doi.org/10.5281/zenodo.6577614} \autocite{sellner_htc_2023}.

\section*{Materials and methods}

\paragraph{Experimental design} 
In this prospective observational study, we collected \ac{hsi} data and corresponding \ac{rgb} images from the skin of patients admitted to the interdisciplinary surgical \ac{icu} at the University Hospital Heidelberg (Germany). All adult patients admitted between October 24, 2022, and December 15, 2023, were included. The study was conducted in accordance with the ethical standards laid down in the 1964 Declaration of Helsinki and its later amendments. The protocol was approved by the Ethics Committee of the Medical Faculty of Heidelberg University, Germany (study reference number: S-288/2022) and registered with the German Clinical Trials Register (study identifier: \href{https://drks.de/search/en/trial/DRKS00029709}{DRKS00029709}) prior to the commencement of recruitment.  The palm and annular finger were selected as measurement sites for their easy accessibility and low melanin content, with the hand chosen to ensure it was not used for intra-arterial cannulas or intravascular access. Structured clinical data were collected alongside the \ac{hsi} data, including demographics, vital signs, blood gas analysis measurements, therapy details (usage of organ replacement therapies, ventilation parameters, dose of administered vasopressors and inotropes), and laboratory results. In total, \varTotalFeaturesTenHrs clinical parameters were recorded, with \varTotalFeaturesOneHrs usually available within one hour and \varTotalFeaturesTenHrs parameters usually available within ten hours of admission. Table~\ref{tab:descriptive_statistics_1hour} and table~\ref{tab:descriptive_statistics_10hours} provide descriptive statistics of the clinical data.

\paragraph{Hyperspectral image acquisition} 
The camera system used was the  medical device-graded TIVITA\textsuperscript{\textregistered} 2.0 Surgery Edition (Diaspective Vision GmbH, Am Salzhaff, Germany). It features a push-broom design with a spectral resolution of approximately \SI{5}{\nm}, covering 100 spectral channels in the range of \SI{500}{\nm} to \SI{1000}{\nm}. The resulting \ac{hsi} cubes have dimensions of $640 \times 480 \times 100$ (width × height × number of spectral channels). The imaged area is approximately $16 \times \SI{11.5}{\cm}$, with an imaging distance of about \SI{50}{cm}, maintained by an integrated distance calibration system. Image acquisition takes approximately \SI{7}{\second}.

The system includes both an \ac{hsi} and \ac{rgb} sensor, providing simultaneous \ac{rgb} images with dimensions of $640 \times 480 \times 3$ (width × height × number of channels). Tissue parameter images, such as oxygen saturation, perfusion index, haemoglobin index, and water index, are estimated from the \ac{hsi} data according to the formulas published in \autocite{holmer_hyperspectral_2018}.

During image acquisition, window blinds were lowered, and all light sources other than the integrated \ac{led} unit were turned off. The hands of patients were supported by the examiner to prevent motion artefacts and ensure more uniform hand positioning, with a consistent background used across all images.

\paragraph{Hyperspectral image annotation} 
Despite using a uniform background and standardising hand positioning as much as possible, images might still include elements such as dressings, wounds, wires, tubes, or parts of the examiner's gloved hand. To mitigate potential confounding from these elements, our analysis was performed on annotated skin areas. We chose circular annotations to consistently capture the same measurement sites across patients, regardless of hand rotation in the imaging plane. According to our annotation guidelines, the selected annotation radii were 100 pixels for the palm and 20 pixels for the ring finger. Finger annotations were centred on the fingertip, and palm annotations were centred on the palm of the hand, defined as the area enclosed by the wrist, the metacarpophalangeal joints and the thumb basal joint.

\paragraph{Sepsis and outcome labels}
Diagnosis of sepsis was based on the Sepsis-3 criteria, which define it as acute, life-threatening organ dysfunction resulting from a suspected or confirmed infection \autocite{singer_sepsis3_2016}. Organ dysfunction was evaluated using the \ac{sofa} score, with an acute increase of at least two points indicating sepsis. Differentiating between organ failure caused by sepsis and that resulting from non-septic inflammation can be challenging, particularly in a surgical \ac{icu} setting following surgical trauma. To maintain label quality and avoid ambiguity in such cases, we introduced a third label, \enquote{unsure}, alongside the labels \enquote{sepsis} and \enquote{no sepsis}. For each patient, the sepsis status was independently assessed by two expert anaesthetists. Disagreements between the two anaesthetists were resolved by a third, more senior anaesthetist (the head of the department for anaesthesia and intensive care). Mortality was assessed through a follow-up conducted 30 days after the patient's inclusion.

\paragraph{Data preprocessing}
Following calibration of \ac{hsi} cubes using white and dark reference cubes, $\ell^1$-normalisation was applied across the spectral channels. The tissue parameter index images oxygen saturation, perfusion index, haemoglobin index, and water index were computed from the \ac{hsi} cubes using the formulas presented in \autocite{holmer_hyperspectral_2018} and subsequently stacked to form a tissue parameter image cube, referred to as TPI cube, with dimensions $640 \times 480 \times 4$ (width $\times$ height $\times$ number of channels). \ac{hsi}, TPI and \ac{rgb} cubes were cropped to a square that tightly encompassed the circular annotation, with pixels outside the annotated area set to zero. The cropped cubes were rescaled to dimensions of $224 \times 224 \times 100$, $224 \times 224 \times 4$ and $224 \times 224 \times 3$, respectively (width $\times$ height $\times$ number of spectral channels). To enable a direct comparison between classification models using \ac{hsi} data from both palm and finger measurement sites versus those using only palm or finger data, cropped \ac{hsi} cubes from both sites for the same patient were stacked along the spectral dimension. This resulted in cubes with dimensions of $224 \times 224 \times 200$, which were used as input for the \model{HSI palm + finger} model.

The missingness in the clinical parameters was low, averaging at \SI{1.6}{\percent}. Missing values were imputed with -1.

\paragraph{Classification models}
We developed deep learning classifiers for automated sepsis diagnosis and mortality prediction using solely \ac{hsi} data (\model{HSI} model), TPI cubes (\model{TPI} model), \ac{rgb} data (\model{RGB} model) and clinical data (\model{clinical data} model), as well as a multimodal approach combining \ac{hsi} with clinical data (\model{HSI + clinical data} model).

The \model{HSI}, \model{TPI} and \model{RGB} models are based on a \ac{cnn} architecture. \Acp{cnn} were chosen for their widespread use in medical \ac{hsi} classification and their advantages over traditional machine learning methods, including higher model accuracy and efficient computation through shared weights and hardware optimizations \autocite{khan_trends_DL}. Using standardised architectures with pretrained weights allows for faster convergence and often yields better performance than training \Acp{cnn} from scratch, particularly when working with small medical datasets \autocite{Tajbakhsh_2016}. To this end, our \model{HSI}, \model{TPI} and \model{RGB} models are composed of a ResNet14d \autocite{he2016deep, rw2019timm} architecture with pre-trained ImageNet weights.

The \model{HSI + clinical data} model consists of two submodels: The \ac{hsi} data are processed equivalently to the \model{HSI} model by a ResNet14d model with pre-trained ImageNet weights up to the bottleneck layer. The clinical data are handled by a submodel comprising two fully connected blocks. Each block includes a linear, batch normalisation, exponential linear unit activation, and dropout layer. The linear layer in the first block has a size of 50, while a size of 30 is used in the second block. The two blocks are followed by a linear head of size ten, matching the bottleneck layer size of the \model{HSI} submodel. After batch normalisation of both bottleneck layers, the bottleneck features are concatenated and fed into another fully connected block, followed by the final classification head.

For the \model{HSI}, \model{TPI}, \model{RGB} and \model{HSI + clinical data} models, cross entropy loss was used during training. The same hyperparameter settings were applied across all deep learning models. Data augmentations included rotations up to $\pm \SI{180}{\degree}$, and horizontal and vertical flipping, each with a probability of 0.5. The AdamW optimizer \autocite{loshchilov2019decoupledweightdecayregularization} and an exponential learning rate schedule were used (initial learning rate: 0.001, decay rate $\gamma$: 0.99, Adam decay rates $\beta_1$: 0.9 and $\beta_2$: 0.999). To regularise the network, a weight decay of 0.001 was applied. Training was conducted for ten epochs, each consisting of 500 images, with stochastic weight averaging \autocite{swa_izmailov2018} applied over the last two epochs. A batch size of 32 images was used, and underrepresented classes were oversampled in each batch to ensure equal class distribution.

For the \model{clinical data} model, we employed a random forest classifier composed of 100 trees, as it is widely used in sepsis diagnosis from \ac{ehr} data \autocite{yang2023predicting}. The implementation from sklearn \autocite{pedregosa-scikit-learn} was used with default settings, except that balanced class weighting was enabled to adjust the weights inversely proportional to the class frequencies in the training data.

\paragraph{Training and validation setup}
The same training and validation setup was used across all trained models. Given the limited dataset size, we decided against a single hold-out test set for model validation. Instead, we implemented a nested cross-validation scheme, providing a more robust performance estimation based on the entire dataset \autocite{varma2006bias}. Both the number of outer and inner folds were set to five.

To further stabilise the performance of the trained networks, each run was repeated three times with different random seed settings, altering the initialization of workers and the order in which images were seen during training. On the validation sets, ensembling was performed over these three repetitions. For the test data, the networks from all five folds and three repetitions each (a total of 15 networks) were ensembled by averaging the predictions (logits).

Following the recommendations in \autocite{metrics_reloaded_maierhein2023}, the model performances were validated using the \ac{roc} curve and \ac{auroc}. To compute confidence intervals that reflect sampling variability, bootstrap sampling was repeated 1000 times for each test set $T$, with $\vert T \vert$ samples randomly drawn with replacement for each bootstrap.

\paragraph{Feature importance of clinical data}
The feature importance of the clinical data was determined using \ac{rfe} \autocite{guyon2002gene} from the \model{clinical data} models. \ac{rfe} was adapted to the 5-fold cross-validation setup of our inner folds by averaging feature importances across inner folds before eliminating the least important feature from the set of input features. This process was performed independently on all five outer folds.

\paragraph{Statistical analysis}
Statistical tests were conducted to identify significant differences in functional tissue parameter values between septic and non-septic patients, as well as between survivors and non-survivors, resulting in four tests for each group. Two-sided Welch’s t-test \autocite{welch1947generalization} was applied, with an overall significance level of 0.05 for each group of tests. To prevent the accumulation of alpha errors due to multiple testing, the Bonferroni correction \autocite{bonferroni1935calcolo} was applied, setting the significance level at 0.0125 per test.


\clearpage 

%
\bibliographystyle{sciencemag}
\bibliography{science_template} 

%
%
%
%
%
%


\section*{Acknowledgments}
During the preparation of this manuscript, the authors used a large language model (GPT-3.5) for language improvements.

\paragraph*{Funding:}
This project has received funding from the European Research Council (ERC) under the European Union’s Horizon 2020 research and innovation programme (project NEURAL SPICING grant agreement No. 101002198) and the National Center for Tumor Diseases (NCT) Heidelberg's Surgical Oncology Program. It was further supported by the German Cancer Research Center (DKFZ) and the Helmholtz Association under the joint research school HIDSS4Health (Helmholtz Information and Data Science School for Health). This publication was supported through state funds approved by the State Parliament of Baden-W\"urttemberg for the Innovation Campus Health + Life Science alliance Heidelberg Mannheim. The European Society of Anaesthesiology and Intensive Care (ESAIC) supported the project with the ESAIC Andreas Hoeft’s Grant Intensive Care in 2023 (Grant ID: ESAIC\_GR\_2023\_MD). Lena Maier-Hein worked with the medical device manufacturer KARL STORZ SE \& Co. KG in the projects \enquote{InnOPlan} and \enquote{OP 4.1}, funded by the German Federal Ministry of Economic Affairs and Energy (grant agreement No. BMWI 01MD15002E and BMWI 01MT17001E) and \enquote{Surgomics}, funded by the German Federal Ministry of Health (grant agreement No. BMG 2520DAT82D).

\paragraph*{Author contributions:}
MD, MAW, LMH and SS conceptualized the study. LMH, MD and MAW acquired the funding and provided resources and supervision. KH and AvG collected and annotated the data. MD, TH, SK, MvdF and MAW labeled the data. SS and JS curated and investigated the data and developed the software and methodology. SS and MD administered the project. SS wrote the manuscript and prepared the visualizations. All authors revised the manuscript.

\paragraph*{Competing interests:}
There are no competing interests to declare.

\paragraph*{Data and materials availability:}
To enable model comparison and external validation, all code and pretrained models will be made publicly available in our GitHub repository (\href{https://github.com/IMSY-DKFZ/htc}{https://github.com/IMSY-DKFZ/htc}) and Zenodo (\href{https://doi.org/10.5281/zenodo.6577614}{https://doi.org/10.5281/zenodo.6577614}) \autocite{sellner_htc_2023}. Patient data, however, cannot be shared due to the absence of consent for data sharing, as approved by the ethics committee.


\subsection*{Supplementary materials}
Figs. S1 to S6\\
Tables S1 to S3\\


\newpage


\renewcommand{\thefigure}{S\arabic{figure}}
\renewcommand{\thetable}{S\arabic{table}}
\renewcommand{\theequation}{S\arabic{equation}}
\renewcommand{\thepage}{S\arabic{page}}
\setcounter{figure}{0}
\setcounter{table}{0}
\setcounter{equation}{0}
\setcounter{page}{1} 


\begin{center}
\section*{Supplementary Materials for\\ \scititle}

Silvia~Seidlitz$^{1,2,3,4\ast}$,
Katharina~Hölzl$^{1,5}$,
Ayca~von~Garrel$^{1,5}$,
Jan~Sellner$^{1,2,3,4}$,\and
Stephan~Katzenschlager$^{5}$,
Tobias~Hölle$^{5}$,
Dania~Fischer$^{5}$,
Maik~von~der~Forst$^{5}$,\and
Felix~C.~F.~Schmitt$^{5}$,
Alexander~Studier-Fischer$^{6,7,8,9}$,
Markus~A.~Weigand$^{5\dagger}$,\and
Lena~Maier-Hein$^{1,2,3,4,10\ast\dagger}$,
Maximilian~Dietrich$^{5\ast\dagger}$
\\

\small$^{1}$Division of Intelligent Medical Systems (IMSY), German Cancer Research Center (DKFZ), Heidelberg, Germany.\\
\small$^{2}$Helmholtz Information and Data Science School for Health, Heidelberg/Karlsruhe, Germany.\\
\small$^{3}$Faculty of Mathematics and Computer Science, Heidelberg University, Heidelberg, Germany.\\
\small$^{4}$National Center for Tumor Diseases (NCT), NCT Heidelberg, a partnership between DKFZ and university\\ \small medical center Heidelberg.\\
\small$^{5}$Heidelberg University, Medical Faculty, Department of Anesthesiology, Heidelberg University Hospital,\\ \small Heidelberg, Germany.\\
\small$^{6}$Department of General, Visceral, and Transplantation Surgery, Heidelberg University Hospital, Heidelberg, Germany.\\
\small$^{7}$Department of Urology and Urosurgery, Medical Faculty of the University of Heidelberg,\\
\small University Medical Center Mannheim, Mannheim, Germany.\\
\small$^{8}$Division of Intelligent Systems and Robotics in Urology (ISRU), German Cancer Research Center (DKFZ),\\ \small Heidelberg, Germany.\\
\small$^{9}$DKFZ Hector Cancer Institute at the University Medical Center Mannheim, Mannheim, Germany.\\
\small$^{10}$Medical Faculty, Heidelberg University, Heidelberg, Germany.\\
\small$^\ast$Corresponding author. Email: s.seidlitz@dkfz-heidelberg.de (S.S.); l.maier-hein@dkfz-heidelberg.de (L.M.-H.); \and \small maximilian.dietrich@med.uni-heidelberg.de (M.D.)\\
\small$^\dagger$These authors contributed equally to this work.
\end{center}

\subsubsection*{This PDF file includes:}
Figures S1 to S6\\
Tables S1 to S3\\

\newpage


\begin{figure}
    \centering
	\includegraphics[width=\textwidth]{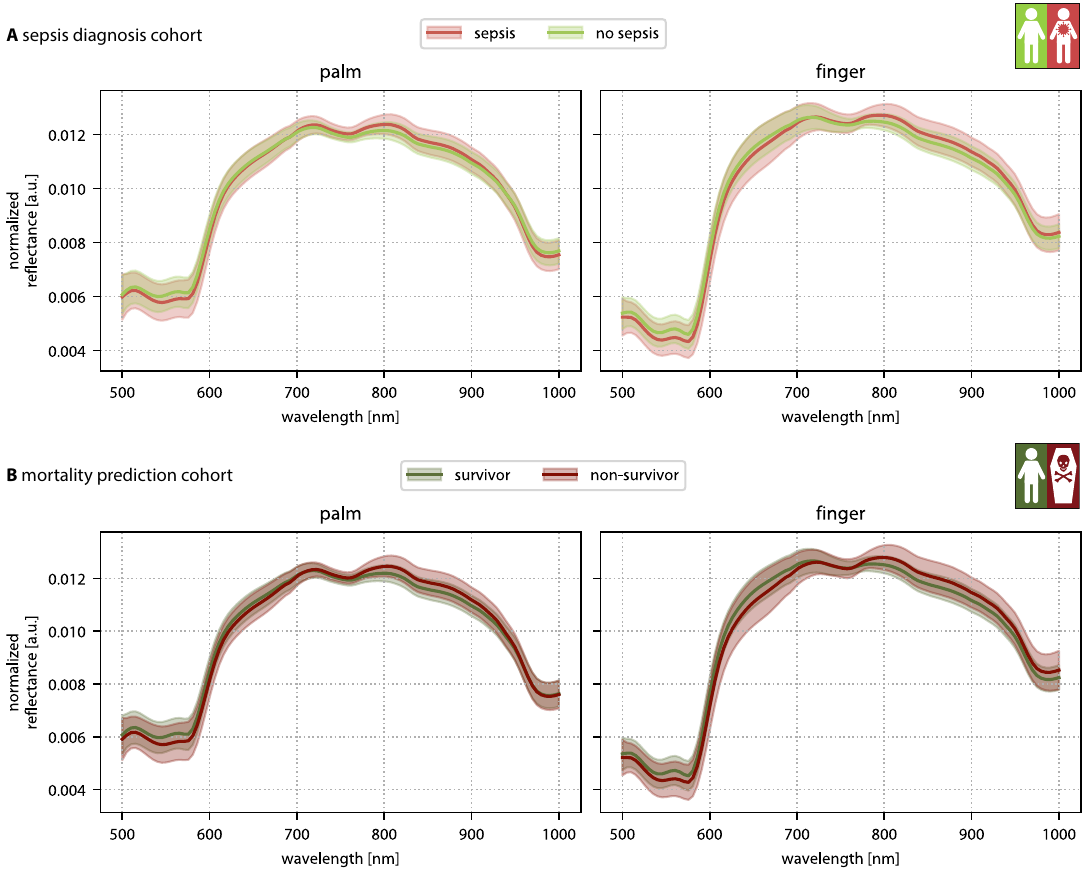}
	\caption{\textbf{Characteristic spectra for septic vs. non-septic patients (A) and survivors vs. non-survivors (B).} The plots display the average $\ell^1$-normalised spectra across patients (solid lines), with shaded areas indicating one standard deviation, for the measurement sites palm (left) and finger (right).
}
	\label{fig:spectra} 
\end{figure}

\begin{figure}
	\centering
	\includegraphics[width=\textwidth]{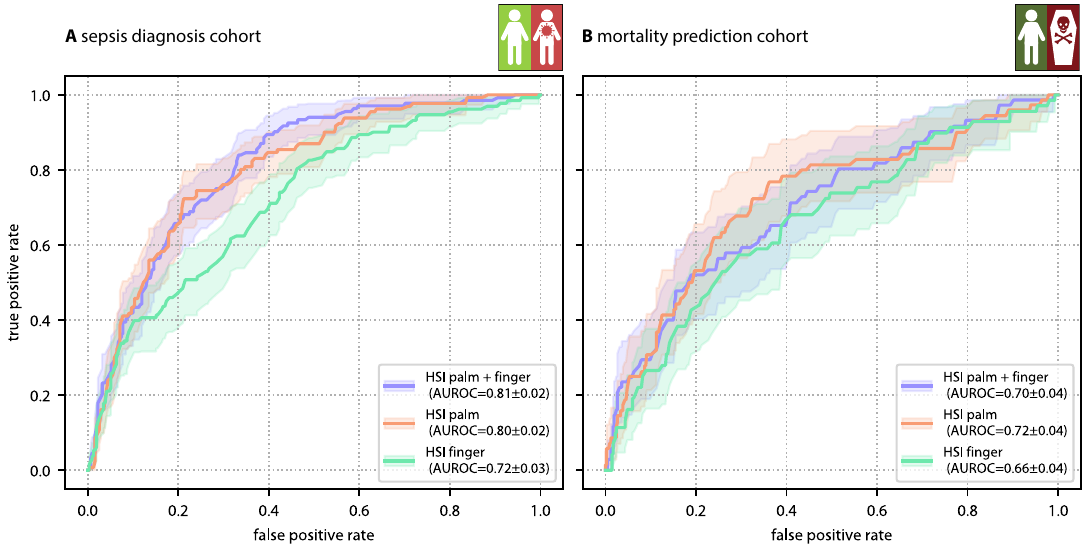}
	\caption{\textbf{Combining palm and finger measurements does not provide a substantial advantage over using palm measurements alone.} \Aclp*{roc} are shown for deep learning-based sepsis diagnosis (\textbf{A}) and mortality prediction (\textbf{B}) using \acf*{hsi} data from the palm (\model{HSI palm} model), the finger (\model{HSI finger} model), and a combination of both (\model{HSI palm + finger} model). Shaded areas represent the \SI{95}{\percent} confidence interval across 1000 bootstrap samples, with the mean and standard deviation of the \acf*{auroc} reported in the legend.}
	\label{fig:performance_combined} 
\end{figure}

\begin{figure}
    \centering
	\includegraphics[width=\textwidth]{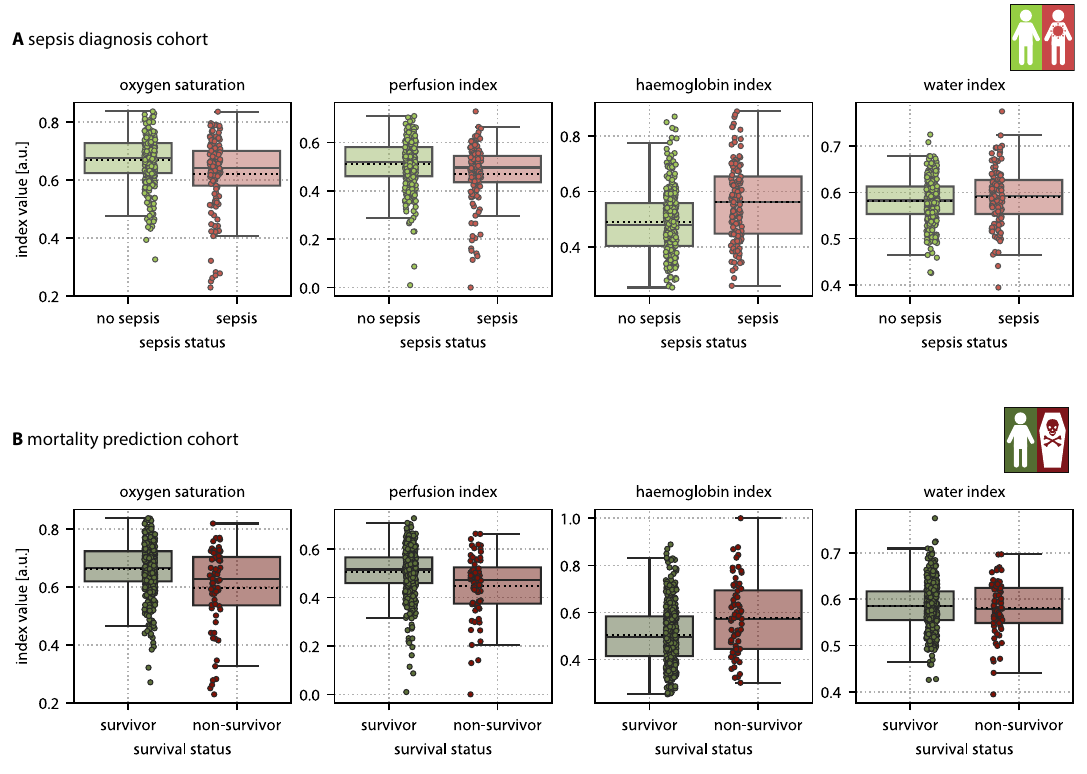}
	\caption{\textbf{Septic patients and non-survivors possess significantly lower finger tissue oxygen saturation, and higher tissue haemoglobin and perfusion index.} The subfigures show the distribution of the functional parameters oxygen saturation, perfusion index, haemoglobin index and water index, derived from \acl*{hsi} finger measurements, for septic and non-septic patients (\textbf{A}), and survivors and non-survivors (\textbf{B}). The boxes denote the quartiles of the distribution with the whiskers extending up to 1.5 times the interquartile range, and the median and mean drawn as solid and dashed lines, respectively. Each dot represents one patient.}
	\label{fig:tpi_finger} 
\end{figure}

\begin{figure}
    \centering
	\includegraphics[width=\textwidth]{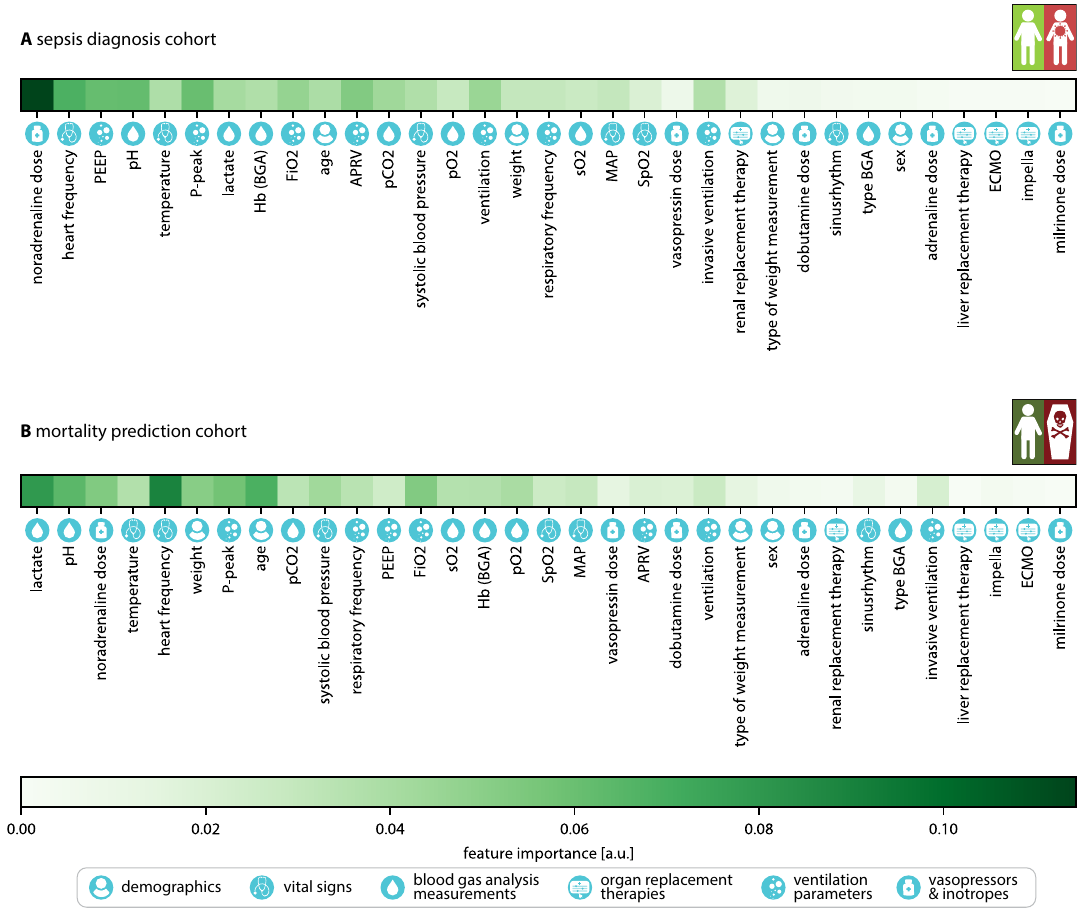}
	\caption{\textbf{Feature importance of clinical data available within one hour of \acl*{icu} admission for sepsis diagnosis (A) and mortality prediction (B) using the \model{clinical data} model.} Colors represent feature importance based on the reduction in Gini importance when a specific feature is used for data splitting within a decision tree node. Clinical data features are ordered by importance as determined through \acl*{rfe} \autocite{guyon2002gene}, from most important (left) to least important (right).}
	\label{fig:feature_importance_1hour} 
\end{figure}

\begin{figure}
    \centering
	\includegraphics[width=\textwidth]{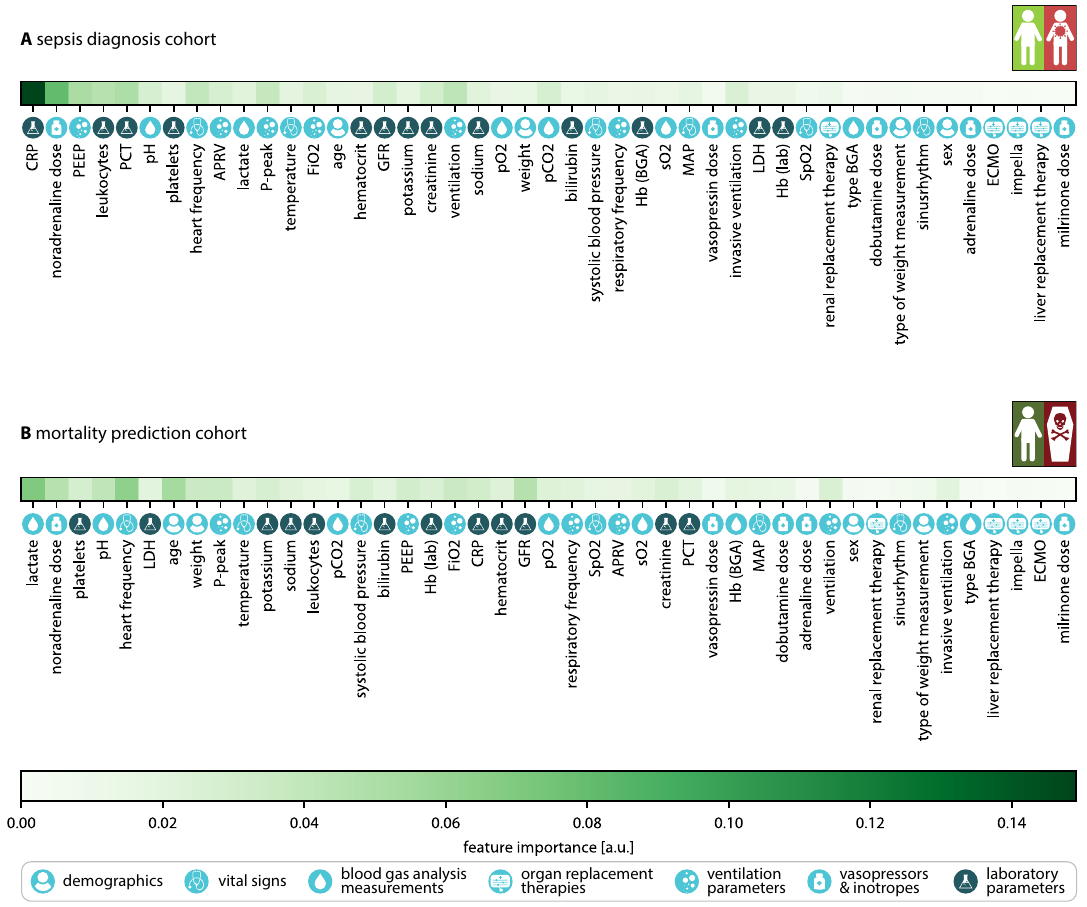}
	\caption{\textbf{Feature importance of clinical data available within ten hours of \acl*{icu} admission for sepsis diagnosis (A) and mortality prediction (B) using the \model{clinical data} model.} Colors represent feature importance based on the reduction in Gini importance when a specific feature is used for data splitting within a decision tree node. Clinical data features are ordered by importance as determined through \acl*{rfe} \autocite{guyon2002gene}, from most important (left) to least important (right).}
	\label{fig:feature_importance_10hours} 
\end{figure}

\begin{figure}
    \centering
	\includegraphics[width=\textwidth]{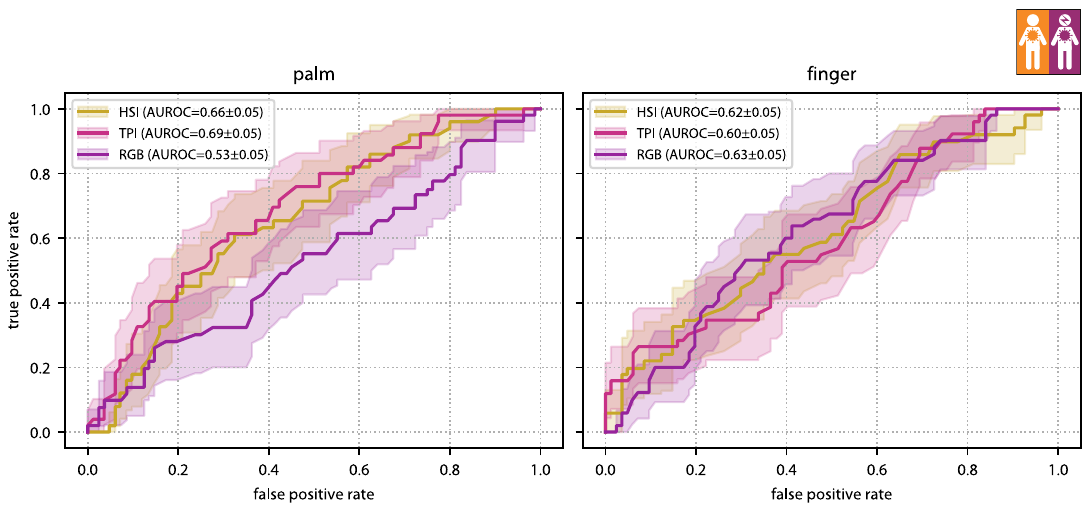}
	\caption{\textbf{\Acf*{hsi} can rapidly and non-invasively differentiate between sepsis and septic shock patients.} Among our \varTotalSepsis septic patients, 49 (\SI{38}{\percent}) experienced septic shock. \Aclp*{roc} are shown for classification models based on \acs*{hsi} data (gold), stacked tissue parameter images (TPI, pink) and \acl*{rgb} (RGB, violet) data of the palm (left) and annular finger (right). The shaded areas denote the \SI{95}{\percent} confidence interval across \num{1000} bootstrap samples, and mean and standard deviation of the \acf*{auroc} are reported in the legend.
}
	\label{fig:shock_performance} 
\end{figure}
\clearpage


\begin{table}
    \centering
    \caption{Two-sided Welch's t-tests \autocite{welch1947generalization}} were performed to determine significant differences in functional tissue parameter values of palm and finger measurements based on sepsis status and survival status. A summary of $p$-values, \acf*{dof}, $t$-statistic and \SI{95}{\percent} \acf*{ci} is provided.
    \label{tab:statistical_testing}
    \varStatisticsTable
\end{table}

\begin{table}[h]
    \fontsize{7.97pt}{7.97pt}\selectfont
    \centering
    \caption{Descriptive statistics are provided for patients with and without sepsis, as well as for survivors and non-survivors. This includes clinical data available within the first hour of admission to the \acl*{icu}, such as demographics, vital signs, \acf*{bga} measurements, use of organ replacement therapies, ventilation parameters, and dose of administered vasopressors and inotropes. For ratio-scaled parameters, means are presented with \acf*{sd} in brackets. For nominal-scaled parameters, the number of patients per category is listed, while for boolean therapy parameters, the percentage of patients receiving the treatment is provided. Abbreviations denote the \acf*{map}, \acf*{spo2}, \acf*{pco2}, \acf*{po2}, \acf*{so2}, \acf*{hb}, \acf*{ecmo}, \acf*{aprv}, \acf*{fio2}, \acf*{peep}, and \acf*{ppeak}.}
    \label{tab:descriptive_statistics_1hour}
    \varDescriptiveTableOneHrs
\end{table}

\begin{table}[h]
    \fontsize{7.97pt}{7.97pt}\selectfont
    \centering
    \caption{Continuation of table~\ref{tab:descriptive_statistics_1hour}, including descriptive statistics for laboratory parameters available within the first ten hours of admission to the \acl*{icu}. Means are presented with \acf*{sd} in brackets. Abbreviations denote the \acf*{gfr}, \acf*{ldh}, \acf*{crp}, \acf*{hb}, and \acf*{pct}}
    \varDescriptiveTableTenHrs
    \label{tab:descriptive_statistics_10hours}
\end{table}

\end{document}